\documentclass[1p]{elsarticle}

\usepackage[numbers]{natbib}
\usepackage{booktabs}
\usepackage{paralist}
\usepackage{verbatim}
\usepackage{url}
\usepackage{hyperref}
\usepackage[inline]{enumitem}
\usepackage[textsize=scriptsize,textwidth=1cm,disable]{todonotes}
\usepackage[xcolor,outerbars]{changebar}

\newenvironment{changed}{%
    \cbcolor{white}%
    \par%
    \cbstart%
}{%
    \cbend%
}

\usepackage{graphicx}
\usepackage{url}
\usepackage{xspace}
\usepackage{comment}
\usepackage{tikz}

\usepackage{pifont}
\newcommand{\yes}{\ding{52}}
\newcommand{\no}{\ding{56}}

\def\citeA{\citet}

\usepackage{amssymb}
\usepackage{xspace}

\def\C{\emph{C}\xspace}
\def\P{\emph{P}\xspace}
\def\Q{\emph{Q}\xspace}

\def\nonevent{\emph{not-}}
\def\notP{\nonevent{\P}\xspace}
\def\notQ{\nonevent{\Q}\xspace}
\def\HP{\emph{H$_\P$}\xspace}
\def\HQ{\emph{H$_\Q$}\xspace}
\def\HNOTQ{\emph{H$_\notQ$}\xspace}
\def\causes{\leadsto}

\usetikzlibrary{arrows,snakes,backgrounds,automata,shapes,patterns,decorations,backgrounds,plotmarks,calc}

\journal{Journal Name}

\begin{document}

\includecomment{short}
\excludecomment{extended}

\begin{frontmatter}

\title{Explanation in Artificial Intelligence:\\ Insights from the Social Sciences}

\author{Tim Miller}

\address{School of Computing and Information Systems\\University of Melbourne, Melbourne, Australia\\\url{tmiller@unimelb.edu.au}}

\begin{abstract}
There has been a recent resurgence in the area of explainable artificial intelligence as researchers and practitioners seek to make their algorithms more understandable. Much of this research is focused on explicitly explaining decisions or actions to a human observer, and it should not be controversial to say that looking at how humans explain to each other can serve as a useful starting point for explanation in artificial intelligence. However, it is fair to say that most work in explainable artificial intelligence uses only the researchers' intuition of what constitutes a `good' explanation. There exists vast and valuable bodies of research in philosophy, psychology, and cognitive science of how people define, generate, select,  evaluate, and present explanations, which argues that people employ certain cognitive biases and social expectations towards the explanation process.  This paper argues that the field of explainable artificial intelligence should build on this existing research, and reviews relevant papers from  philosophy, cognitive psychology/science, and social psychology, which study these topics. It draws out some important findings, and discusses ways that these can be infused with work on explainable artificial intelligence.
\end{abstract}

\begin{keyword}
Explanation \sep Explainability \sep Interpretability \sep Explainable AI \sep Transparency

\end{keyword}

\end{frontmatter}





\tableofcontents

\section{Introduction}

Recently, the notion of \emph{explainable artificial intelligence} has seen a resurgence, after having slowed since the burst of work on explanation in expert systems over three decades ago; for example, see \citeA{chandrasekaran1989explaining}, \cite{swartout1993explanation}, and \citeA{buchanan1984rule}. Sometimes abbreviated XAI (eXplainable artificial intelligence), the idea can be found in grant solicitations~\cite{DARPA2016}
and in the popular press~\cite{Nott17}.  This resurgence is driven by evidence that many AI applications have limited take up, or are not appropriated at all, due to ethical concerns \cite{angwin2016machine} and a \emph{lack of trust} on behalf of their users \cite{stubbs2007autonomy,linegang2006human}. The running hypothesis is that by building more transparent, interpretable, or explainable systems, users will be better equipped to understand and therefore trust the intelligent agents \cite{mercado2016intelligent,chen2014situation,hayes2017improving}.

While there are many ways to increase trust and transparency of intelligent agents, two complementary approaches will form part of many trusted autonomous systems: (1) generating decisions\footnote{We will use \emph{decision}  as the general term to encompass outputs from AI systems, such as categorisations, action selection, etc.} in which one of the criteria taken into account during the computation is how well a human could understand the decisions in the given context, which is often called \emph{interpretability} or \emph{explainability}; and (2) explicitly explaining decisions to people, which we will call \emph{explanation}. Applications of explanation are considered in many sub-fields of artificial intelligence, such as justifying autonomous agent behaviour \cite{mercado2016intelligent,hayes2017improving}, debugging of machine learning models \cite{kulesza2015principles}, explaining medical decision-making \cite{fox2007argumentation}, and explaining predictions of classifiers \cite{ribeiro2016should}.

If we want to design, and implement intelligent agents that are truly capable of providing explanations to \emph{people}, then it is fair to say that models of how humans explain decisions and behaviour to each other are a good way to start analysing the problem.  Researchers argue that people employ certain \emph{biases} \cite{kahneman2011thinking} and \emph{social expectations} \cite{hilton1990conversational} when they generate and evaluate explanation, and I argue that such biases and expectations can improve human interactions with explanatory AI. For example,
\citeA{degraaf2017people} argues that because people assign human-like traits to artificial agents, people will expect explanations using the same conceptual framework used to explain human behaviours.

 Despite the recent resurgence of explainable AI, most of the research and practice in this area seems to use the researchers' intuitions of what constitutes a `good' explanation. \citeA{miller2017inmates} shows in a small sample that research in explainable AI typically does not cite or build on frameworks of explanation from social science. They argue that this could lead to failure. The very experts who understand decision-making models the best are not in the right position to judge the usefulness of explanations to lay users --- a phenomenon that \citeauthor{miller2017inmates} refer to (paraphrasing \citeA{cooper2004inmates}) as ``the inmates running the asylum''. Therefore, a strong understanding of how people define, generate, select, evaluate, and present explanations seems almost essential.

In the fields of philosophy, psychology, and cognitive science, there is a vast and mature body of work that studies these exact topics. For millennia, philosophers have asked the questions about what constitutes an explanation, what is the function of explanations, and what are their structure. For over 50 years, cognitive and social psychologists have analysed how people attribute and evaluate the social behaviour of others. For over two decades, cognitive psychologists and scientists have investigated how people generate explanations and how they evaluate their quality. 

I argue here that there is considerable scope to infuse this valuable body of research into explainable AI. Building intelligent agents capable of explanation is a challenging task, and approaching this challenge in a vacuum considering only the computational problems will not solve the greater problems of trust in AI. Further, while some recent work builds on the early findings on explanation in expert systems, that early research was undertaken prior to much of the work on explanation in social science. I contend that newer theories can form the basis of explainable AI --- although there is still a lot to learn from early work in explainable AI around design and implementation.



This paper aims to promote the inclusion of this existing research into the field of explanation in AI. As part of this work, over 250 publications on explanation were surveyed from social science venues. A smaller subset of these were chosen to be presented in this paper, based on their currency and relevance to the topic. The paper presents relevant theories on explanation, describes, in many cases, the experimental evidence supporting these theories, and presents ideas on how this work can be infused into explainable AI. 

\subsection{Scope}
\label{sec:intro:scope}

\begin{changed}
In this article, the term `\emph{Explainable AI}' loosely refers to an explanatory agent revealing underlying causes to its or another agent's decision making. However, it is important to note that the solution to explainable AI is not just `more AI'. Ultimately, it is a human-agent interaction problem. Human-agent interaction can be defined as the intersection of artificial intelligence, social science, and human-computer interaction (HCI); see Figure~\ref{fig:XAI-scope}. Explainable AI is just one problem inside human-agent interaction. 

\begin{figure}[!t]
\centering
\scalebox{1.0}[1.0]{
\begin{tikzpicture}
{\small
  \tikzset{venn circle/.style={draw,circle,minimum width=6cm,fill=#1,opacity=0.4}}

  \node [venn circle = gray] (AI) at (0,0) {};
  \node [venn circle = gray] (SS) at (60:3cm) {};
  \node [venn circle = gray] (HCI) at (0:3cm) {};
  \node[above,xshift=-13mm,yshift=-5mm] at (barycentric cs:AI=1 ) {Artificial};
  \node[below,xshift=-13mm,yshift=-4mm] at (barycentric cs:AI=1 ) {Intelligence};
  \node[above,yshift=14mm] at (barycentric cs:SS=1 ) {Social};
  \node[below,yshift=15mm] at (barycentric cs:SS=1 ) {Science};
  \node[above,xshift=15mm, yshift=-5mm] at (barycentric cs:HCI=1 ) {Human-Computer};
  \node[below,xshift=15mm, yshift=-4mm] at (barycentric cs:HCI=1 ) {Interaction};
  \node[above,yshift=-3mm] at (barycentric cs:AI=1/3,SS=1/3,HCI=1/3 ){Human-Agent};
  \node[below,yshift=-2mm] at (barycentric cs:AI=1/3,SS=1/3,HCI=1/3 ){Interaction};

  \node [venn circle = gray, minimum width=0.5cm,opacity=0.7,yshift=8mm] (XAI) at (barycentric cs:AI=1/3,SS=1/3,HCI=1/3 ) {XAI};
}
\end{tikzpicture}
}
\caption{Scope of Explainable Artificial Intelligence}
\label{fig:XAI-scope}
\end{figure}
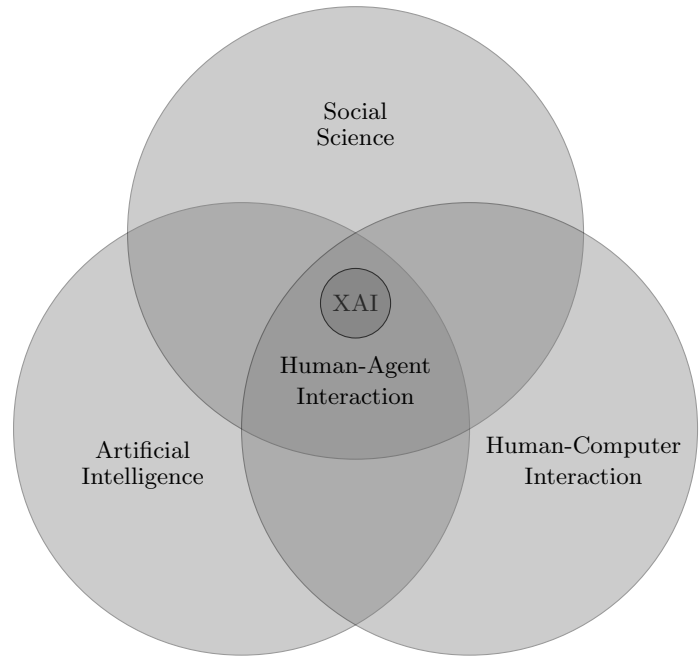

This article highlights the top circle in Figure~\ref{fig:XAI-scope}: the philosophy, social and cognitive psychology, and cognitive science views of explanation, and their relation to the other two circles: their impact on the design of both artificial intelligence and our interactions with them. With this scope of explainable AI in mind, the scope of this article is threefold:
\end{changed}

\begin{itemize}
 \item \emph{Survey}: To survey and review relevant articles on the philosophical, cognitive, and social foundations of explanation, with an emphasis on `everyday' explanation.
 
 \item \emph{Everyday explanation}: To focus on `everyday' (or local) explanations as a tool and process for an agent, who we call the \emph{explainer}, to explain decisions made by \emph{itself or another agent} to a \emph{person}, who we call the \emph{explainee}. 
%
 `Everyday' explanations are the explanations of why particular facts (events, properties, decisions, etc.) occurred, rather than explanations of more general relationships, such as those seen in scientific explanation. We justify this focus  based on the observation from AI literature that trust is lost when users cannot understand traces of observed behaviour or decisions \cite{stubbs2007autonomy,mercado2016intelligent}, rather than trying to understand and construct generalised theories. Despite this, everyday explanations also sometimes refer to generalised theories, as we will see later in Section~\ref{sec:philosophical-foundations}, so scientific explanation is relevant, and some work from this area is surveyed in the paper.
 
 \item \emph{Relationship to Explainable AI}: To draw important points from relevant articles to some of the different sub-fields of explainable AI.

 \end{itemize}
 
\noindent The following topics are considered \emph{out of scope} of this article:
 
 \begin{itemize}
 \item \emph{Causality}: While causality is important in explanation, this paper is not a survey on the vast work on causality. I review the major positions in this field insofar as they relate to the relationship with models of explanation.

 
 \item \emph{Explainable AI}: This paper is not a survey on existing approaches to explanation or interpretability in AI, except those that directly contribute to the topics in scope or build on social science. For an excellent short survey on explanation in machine learning, see \citeA{biran2017explanation}.


\end{itemize}

\subsection{Major Findings}

As part of this review, I highlight four major findings from the surveyed literature that I believe are important for explainable AI, but which I believe most research and practitioners in artificial intelligence are currently unaware:

\begin{enumerate}

 \item Explanations are \emph{contrastive} --- they are sought in response to particular \emph{counterfactual cases}, which are termed \emph{foils} in this paper. That is, people do not ask why event \P happened, but rather why event \P happened \emph{instead of} some event \Q. This has important social and computational consequences for explainable AI. In Sections~\ref{sec:philosophical-foundations}--\ref{sec:cognitive-processes}, models of how people provide contrastive explanations are reviewed.
\begin{changed}
 \item Explanation are \emph{selected} (in a biased manner) --- people rarely, if ever, expect an explanation that consists of an actual and complete cause of an event. Humans are adept at selecting one or two causes from a sometimes infinite number of causes to be \emph{the} explanation. However, this selection is influenced by certain cognitive biases. In Section~\ref{sec:cognitive-processes},  models of how people select explanations, including how this relates to contrast cases, are reviewed.
\end{changed}
\begin{changed}
 \item Probabilities probably don't matter --- while truth and likelihood are important in explanation and probabilities really do matter, \emph{referring} to probabilities or statistical relationships in explanation is not as effective as referring to causes. The most likely explanation is not always the \emph{best} explanation for a person, and importantly, using statistical generalisations to explain why events occur is unsatisfying, unless accompanied by an underlying \emph{causal} explanation for the generalisation itself.
\end{changed}
 \item Explanations are \emph{social} --- they are a transfer of knowledge, presented as part of a conversation\footnote{Note that this does not imply that explanations must be given in natural language, but implies that explanation is a social interaction between the explainer and the explainee.} or interaction, and are thus presented relative to the explainer's beliefs about the explainee's beliefs. In Section~\ref{sec:social-explanation},  models of how people interact regarding explanations are reviewed.

\end{enumerate}

These four points all converge around a single point: explanations are not just the presentation of associations and causes (\emph{causal attribution}), they are \emph{contextual}. While an event may have many causes, often the explainee cares only about a small subset (relevant to the context), the explainer selects a subset of this subset (based on several different criteria), and explainer and explainee may interact and argue about this explanation.

I assert that, if we are to build truly explainable AI, especially intelligent systems that are able to offer explanations, then these three points are imperative in many applications.

\subsection{Outline}

The outline of this paper is as follows. Section~\ref{sec:intro:example} presents a motivating example of an explanatory agent that is used throughout the paper. Section~\ref{sec:philosophical-foundations} presents the philosophical foundations of explanation, defining what explanations are, what they are not, how to relate to causes, their meaning and their structure. Section~\ref{sec:social-attribution} focuses on one specific type of explanation --- those relating to human or social behaviour, while Section~\ref{sec:cognitive-processes} surveys work on how people generate and evaluate explanations more generally; that is, not just social behaviour. Section~\ref{sec:social-explanation} describes research on the dynamics of interaction in explanation between explainer and explainee. Section~\ref{sec:conc} concludes and highlights several major challenges to explanation in AI.

\subsection{Example}
\label{sec:intro:example}

This section presents a simple example, which is used to illustrate many important concepts through this paper. It is of a hypothetical system that categorises images of arthropods into several different types, based on certain physical features of the arthropods, such as number of legs, number of eyes, number of wings, etc.  The algorithm is assumed to have been trained on a large set of valid data and is highly accurate. It is used by entomologists to do automatic classification of their research data.  
Table~\ref{tab:arthropod} outlines a simple model of the features of arthropods for illustrative purposes.
An explanation function is available for the arthropod system.

\begin{table}[!th]
\centering
\begin{tabular}{lcccccc}
\toprule
 &  & & & \textbf{Compound} & \\[-1mm]
\textbf{Type} & \textbf{No.\ Legs} & \textbf{Stinger} & \textbf{No.\ Eyes} & \textbf{Eyes} & \textbf{Wings}\\

\midrule
Spider & 8 & \no  & 8 & \no  & 0\\
Beetle & 6 & \no  & 2 & \yes  & 2\\
Bee    & 6 & \yes & 5 & \yes  & 4\\
Fly    & 6 & \no  & 5 & \yes & 2\\
\bottomrule
\end{tabular}
\caption{A simple lay model for distinguishing common arthropods.}
\label{tab:arthropod}
\end{table}

Now, consider the idealised and simple dialogue between a human user and `ExplAgent', who is the interactive explanation agent, outlined in Table~\ref{fig:example-dialogue}. This dialogue is not intended to be realistic,  but is merely illustrative of how a particular explanatory agent may interact: responding to posed questions, using mixed modalities --- in this case, language and visual images --- and being able to answer a range of questions about its decision making. This example shows different types of questions being posed, and demonstrates that the explanatory agent will need to keep track of the state of the explanation; for example, by noting what it has already told the explainee, and may have to infer what the explainee has inferred themselves.

\begin{figure}[!th]
\centering
\begin{tabular}{p{0.12\textwidth}p{0.8\textwidth}}
\toprule
Person: & ``\emph{Why is image \emph{J} labelled as a \emph{Spider} instead of a \emph{Beetle}?}''\\

ExplAgent: & ``\emph{Because the arthropod in image \emph{J} has eight legs, consistent with those in the category \emph{Spider}, while those in \emph{Beetle} have six legs.}''\\

Person: & ``\emph{Why did you infer that the arthropod in image \emph{J} had eight legs instead of six?}'' \\

ExplAgent: &  ``\emph{I counted the eight legs that I found, as I have just highlighted on the image now.}'' (ExplAgent shows the image with the eight legs counted).\\

Person: & ``\emph{How do you know that spiders have eight legs?}''\\

ExplAgent: & ``\emph{Because in the training set I was trained on, almost all animals with eight legs were labelled as \emph{Spider}.}''\\

Person: &  ``\emph{But an octopus can have eight legs too. Why did you not classify image J as an octopus?}''\\

ExplAgent: & ``\emph{Because my function is only to classify arthropods.}''\\
\bottomrule
\end{tabular}
\caption{Example Explanation Dialogue between a Person and an Explanation Agent}
\label{fig:example-dialogue}
\end{figure}

We will refer back to this example throughout the paper and link difference parts of work the different parts of the dialogue above.


\section{Philosophical Foundations --- What Is Explanation?}
\label{sec:philosophical-foundations}

\begin{quote} 
To explain an event is to provide some information about its causal history. In an act of explaining, someone who is in possession of some information  about the causal history of some event --- \emph{explanatory information}, I shall call it --- tries to convey it to someone else. -- \citeA[p.\ 217]{lewis1986causal}
\end{quote}

In this section, we outline foundational work in explanation, which helps to define causal explanation and how it differs from other concepts such as causal attribution and interpretability.

\subsection{Definitions}

There are several related concepts in explanation, which seem to be used interchangeably between authors and also within articles, often demonstrating some conflation of the terms. In particular, this section describes the difference between causal attribution and causal explanation. 
We will also briefly touch on the difference between explanation and interpretability.

\subsubsection{Causality}
\label{sec:philosophical-foundations:causality}

The idea of causality has attracted much work, and there are several different accounts of what constitutes a \emph{cause} of an event or property. The various definitions of causation can be broken into two major categories: \emph{dependence theories} and \emph{transference theories}.

\paragraph{Causality and Counterfactuals}
Hume \cite[Section~VII]{hume2000enquiry} is credited with deriving what is known as the \emph{regularity theory} of causation. This theory states that there is a cause between two types of events if events of the first type are always followed by events of the second. However, as argued by \citeA{lewis1974causation}, the definition due to \citeauthor{hume2000enquiry} is in fact about \emph{counterfactuals}, rather than dependence alone. \citeauthor{hume2000enquiry}  argues that the co-occurrence of events $C$ and $E$, observed from experience, do not give causal information that is useful.  Instead, the cause should be understood relative to an imagined, counterfactual case: event $C$ is said to have caused event $E$ if, under some hypothetical counterfactual case the event $C$ did not occur, $E$ would not have occurred. This definition has been argued and refined, and many definitions of causality are based around this idea in one way or another; c.f.\ \citeA{lewis1974causation,hilton1988logic}.

This \emph{classical counterfactual} model of causality is well understood but competing definitions exist. \emph{Interventionist} theories of causality \cite{woodward2005making,halpern2005causes-part-I} state that event $C$ can be deemed a cause of event $E$ if and only if any change to event $E$ can be brought about solely by intervening on event $C$. \emph{Probabilistic} theories, which are extensions of interventionist theories, state that event $C$ is a cause of event $E$ if and only if the occurrence of $C$ increases the probability of $E$ occurring \cite{menzies1993causation}.

 \emph{Transference} theories \cite{aronson1971grammar,fair1979causation,dowe1992wesley}, on the other hand, are not defined on dependence, but instead describe \emph{physical} causation as the transference of energy between objects. In short, if $E$ is an event representing the change of energy of an object $O$, then $C$ causes $E$ if object $O$ is in contact with the object that causes $C$, and there is some \emph{quantity} of energy transferred.

While the aim here is not a detailed survey of causality, however, it is pertinent to note that the dependence theories all focus around the concept of \emph{counterfactuals}: the state of affairs that \emph{would have resulted} from some event that \emph{did not occur}. Even transference theories, which are not explicitly defined as counterfactual, consider that causation is an \emph{unnatural} transference of energy to the receiving object, implying what would have been otherwise. As such, the notion of `counterfactual' is important in causality.

\citeA{gerstenberg2017eye} tested whether people consider counterfactuals when making causal judgements in an experiment involving colliding balls. They presented experiment participants with different scenarios involving two balls colliding, with each scenario having different outcomes, such as one ball going through a gate, just missing the gate, or missing the gate by a long distance.  While wearing eye-tracking equipment, participants were asked to determine what the outcome would have been (a counterfactual) had the candidate cause not occurred (the balls had not collided). Using the eye-gaze data from the tracking, they showed that their participants, even in these physical environments, would trace where the ball would have gone had the balls not collided, thus demonstrating that they used counterfactual simulation to make causal judgements.

\paragraph{Necessary and Sufficient Causes}
\citeA{kelley1972causal} proposes a taxonomy of causality in social attribution, but which has more general applicability, and noted that there are two main types of \emph{causal schemata} for causing events: \emph{multiple necessary causes} and \emph{multiple sufficient causes}. The former defines a schema in which a set of events are all necessary to cause the event in question, while the latter defines a schema in which there are multiple possible ways to cause the event, and only one of these is required. Clearly, these can be interleaved; e.g.\ causes $C_1$, $C_2$, and $C_3$ for event $E$, in which $C_1$ is necessary and either of $C_2$ or $C_3$ are necessary, while both $C_2$ and $C_3$ are sufficient to cause the compound event $(C_2~or~C_3)$.


\paragraph{Internal and External Causes} 
\citeA{heider1958psychology}, the grandfather of causal attribution in social psychology, argues that causes fall into two camps: internal and external. Internal causes of events are those due to the characteristics of an actor, while external causes are those due to the specific situation or the environment. Clearly, events can have causes that mix both. However, the focus of work from \citeauthor{heider1958psychology} was not on causality in general, but on \emph{social attribution}, or the \emph{perceived} causes of behaviour. That is, how people attribute the behaviour of others. Nonetheless, work in this field, as we will see in Section~\ref{sec:social-attribution}, builds heavily on counterfactual causality.

\paragraph{Causal Chains}
In causality and explanation, the concept of \emph{causal chains} is important. A causal chain is a path of causes between a set of events, in which a cause from event $C$ to event $E$ indicates that $C$ must occur before $E$. Any events without a cause are \emph{root causes}.

\citeA{hilton2005course} define five different types of causal chain, outlined in Table~\ref{tab:cognitive-processes:causal-chain}, and note that different causal chains are associated with different types of explanations.

\begin{table}[!ht]
\centering 
\begin{small}
\begin{tabular}{p{0.15\textwidth}p{0.4\textwidth}p{0.4\textwidth}}
\toprule
 \textbf{Type} & \textbf{Description} & \textbf{Example}\\
 \midrule
 Temporal & Distal events do not constraint proximal events. Events can be switched in time without changing the outcome & \emph{A} and \emph{B} together cause \emph{C}; order of \emph{A} and \emph{B} is irrelevant; e.g.\ two people each flipping a coin win if both coins are heads; it is irrelevant who flips first.\\
 Coincidental & Distal events do not constraint proximal events. The causal relationships holds in a particular case, but not in general. & \emph{A} causes \emph{B} this time, but the general relationship does not hold; e.g.\ a person smoking a cigarette causes a house fire, but this does not generally happen.\\
 Unfolding & Distal events strongly constrain proximal events. The causal relationships hold in general and in this particular case and cannot be switched. & \emph{A} causes \emph{B} and \emph{B} causes \emph{C}; e.g.\ switching a light switch causes an electric current to run to the light, which causes the light to turn on\\
 Opportunity chains & The distal event \emph{enables} the proximal event.  & \emph{A} enables \emph{B}, \emph{B} causes \emph{C}; e.g.\ installing a light switch enables it to be switched, which causes the light to turn on.\\
 Pre-emptive & Distal precedes proximal and prevents the proximal from causing an event. & \emph{B} causes \emph{C}, \emph{A} would have caused \emph{C} if \emph{B} did not occur; e.g.\ my action of unlocking the car with my remote lock would have unlocked the door if my wife had not already unlocked it with the key.\\
\bottomrule
\end{tabular}
\end{small}
\caption{Types of Causal Chains according to \citeA{hilton2005course}.}
\label{tab:cognitive-processes:causal-chain}
\end{table}

People do not need to understand a complete causal chain to provide a sound explanation. This is evidently true: causes of physical events can refer back to events that occurred during the Big Bang, but nonetheless, most adults can explain to a child why a bouncing ball eventually stops. 

\paragraph{Formal Models of Causation}
While  several formal models of causation have been proposed, such as those based on conditional logic \cite{giordano2004conditional, lewis1974causation}, the model of causation that I believe would be of interest to many in artificial intelligence is the formalisation of causality by \citeA{halpern2005causes-part-I}. This is a general model that should be accessible to anyone with a computer science background, has been adopted by philosophers and psychologists, and is accompanied by many additional results, such as an axiomatisation \cite{halpern2000axiomatizing} and a series articles on complexity analysis \cite{eiter2002complexity,eiter2006causes}.

\citeA{halpern2005causes-part-I} define a model-based approach using \emph{structural causal models}  over  two sets of variables: \emph{exogenous} variables, whose values are determined by factors external to the model, and \emph{endogenous} variables, whose values are determined by relationships with other (exogenous or endogenous) variables. Each endogenous variable has a function that defines its value from other variables. A \emph{context} is an assignment of values to variables. Intuitively, a context represents a `possible world' of the model. A model/context pair is called a \emph{situation}. Given this structure,  \citeauthor{halpern2005causes-part-I} define a \emph{actual cause} of an event $X=x$ (that is, endogenous variable $X$ receiving the value $x$) as a set of events $E$ (each of the form $Y=y$) such that (informally) the following three criteria hold:

\begin{description}

 \item [AC1] Both the event $X=x$ and the cause $E$ are true in the actual situation.

 \item [AC2] If there was some \emph{counterfactual} values for the variables of the events in $E$, then the event $X=x$ would not have occurred.

 \item [AC3] $E$ is minimal --- that is, there are no irrelevant events in the case.

\end{description}

A \emph{sufficient cause} is simply a non-minimal actual cause; that is, it satisfies the first two items above.

We will return later to this model in Section~\ref{sec:social-explanation:relevance} to to discuss \citeauthor{halpern2005causes-part-I}'s model of explanation.

\subsubsection{Explanation}
\label{sec:philosophical:definitions:explanation}

\begin{quote}
An explanation is an assignment of causal responsibility --- \citeA{josephson1996abductive}
\end{quote}

Explanation is both a process and a product, as noted by \citeA{lombrozo2006structure}. However, I argue that there are actually \emph{two} processes in explanation, as well as the product:

\begin{enumerate}

 \item \emph{Cognitive process} --- The process of abductive inference for `filling the gaps' \cite{chin2010background} to  determine  an explanation for a given event, called the \emph{explanandum}, in which the causes for the event are identified, perhaps in relation to a particular counterfactual cases, and a subset of these causes is selected as \emph{the explanation} (or \emph{explanans}).

In social science, the process of identifying the causes of a particular phenomenon is known as \emph{attribution}, and is seen as just \emph{part} of the entire process of explanation.

 \item \emph{Product} --- The explanation that results from the cognitive process is the \emph{product} of the cognitive explanation process. 

 \item \emph{Social process} --- The process of transferring knowledge between explainer and explainee, generally an interaction between a group of people, in which the goal is that the explainee has enough information to understand the \emph{causes} of the event; although other types of goal exists, as we discuss later.

\end{enumerate}


But what constitutes an explanation? This question has created a lot of debate in philosophy, but accounts of explanation both philosophical and psychology stress the importance of causality in explanation --- that is, an explanation refers to causes \cite{salmon2006four,woodward2005making,lombrozo2010causal,halpern2005causes-part-II}. There are, however, definitions of non-causal explanation \cite{ginet2008defense}, such as explaining `what happened' or explaining what was meant by a particular remark \cite{wendt1998constitution}. These definitions out of scope in this paper, and they present a different set of challenges to explainable AI.

\subsubsection{Explanation as a Product}

We take the definition that an explanation is an answer to a \emph{why--question} \cite{dennett1989intentional,overton2011scientific,lewis1986causal,lipton1990contrastive}.

According to \citeA{bromberger1966whyquestions}, a why-question is a combination of a \emph{whether--question}, preceded by the word `why'. A whether-question is an interrogative question whose correct answer is either `yes' or `no'.  The \emph{presupposition} within a why--question is the fact referred to in the question that is under explanation, expressed as if it were true (or false if the question is a negative sentence). For example, the question ``\emph{why did they do that?}'' is a why-question, with the inner whether-question being ``\emph{did they do that?}'', and the presupposition being ``\emph{they did that}''. However, as we will see in Section~\ref{sec:philosophical-foundations:contrastive-explanation}, why--questions are structurally more complicated than this: they are \emph{contrastive}.


\begin{changed}
\begin{table}[!t]
\centering
\begin{tabular}{llp{8.2cm}}
\toprule
\textbf{Question} & \textbf{Reasoning} & \textbf{Description}\\
\midrule
What? &  Associative & Reason about which unobserved events could have occurred given the observed events\\[2mm]
How? & Interventionist & Simulate a change in the situation to see if the event still happens \\[2mm]
Why? & Counterfactual & Simulating alternative causes to see whether the event still happens \\
\bottomrule
\end{tabular}
\caption{Classes of Explanatory Question and the Reasoning Required to Answer}
\label{tab:explanatory-questions}
\end{table}

However, other types of questions can be answered by explanations.
In Table~\ref{tab:explanatory-questions}, I propose a simple model for explanatory questions based on \citeauthor{pearl2018book}'s \emph{Ladder of Causation} \cite{pearl2018book}. This model places explanatory questions into three classes: (1) \emph{what}--questions, such as ``\emph{What event happened?}''; (2) \emph{how}-questions, such as ``\emph{How did that event happen?}''; and (3) \emph{why}--questions, such as ``\emph{Why did that event happen?}''. From the perspective of reasoning, \emph{why}--questions are the most challenging, because they use the most sophisticated reasoning. \emph{What}-questions ask for factual accounts, possibly using associative reasoning to determine, from the observed events, which unobserved events also happened. \emph{How} questions are also factual, but require interventionist reasoning to determine the set of causes that, if removed, would prevent the event from happening. This may also require associative reasoning. We categorise \emph{what if}--questions has \emph{how}--questions, as they are just a contrast case analysing what would happen under a different situation.  \emph{Why}--questions are the most challenging, as they require counterfactual reasoning to undo events and simulate other events that are not factual. This also requires associative and interventionist reasoning.

\citeA{dennett2017bacteria} argues that ``why'' is ambiguous and that there are two different senses of why--question: \emph{how come?} and \emph{what for?}. The former asks for a \emph{process narrative}, without an explanation of what it is \emph{for}, while the latter asks for a \emph{reason}, which implies some intentional thought behind the cause. \citeauthor{dennett2017bacteria} gives the examples of ``why are planets spherical?'' and ``why are ball bearings spherical?''. The former asks for an explanation based on physics and chemistry, and is thus a how-come--question, because planets are not round \emph{for} any reason. The latter asks for an explanation that gives the reason what the designer made ball bearings spherical \emph{for}:  a reason because people design them that way.
\end{changed}

Given a why--question, \citeA{overton2011scientific} defines an \emph{explanation} as a pair consisting of: (1) the \emph{explanans}: which is the answer to the question; and (2) and the \emph{explanandum}; which is the presupposition.

\begin{changed}
\subsubsection{Explanation as Abductive Reasoning}
\label{sec:philosophical-foundations:explanation:explanation-as-process}

As a cognitive process, explanation is closely related to \emph{abductive reasoning}. \citeA{peirce1903pragmatism} was the first author to consider abduction as a distinct form of reasoning, separate from induction and deduction, but which, like induction, went from effect to cause. His work focused on the difference between accepting a hypothesis via scientific experiments (induction), and \emph{deriving} a hypothesis to explain observed phenomenon (abduction). He defines the form of inference used in abduction as follows:

\begin{quote}
The surprising fact, $C$, is observed;\\
But if $A$ were true, $C$ would be a matter of course,\\
Hence, there is reason to suspect that $A$ is true.
\end{quote}

Clearly, this is an inference to \emph{explain} the fact $C$ from the hypothesis $A$, which is different from deduction and induction. However, this does not account for competing hypotheses. \citeA{josephson1996abductive} describe this more competitive-form of abduction as:

\begin{quote}
$D$ is a collection of data (facts, observations, givens).\\
$H$ explains $D$ (would, if true, explain $D$).\\
No other hypothesis can explain $D$ as well as $H$ does.\\
Therefore, $H$ is probably true.
\end{quote}

\citeA{harman1965inference} labels this process ``\emph{inference to the best explanation}''. Thus, one can think of abductive reasoning as the following process: (1) observe some (presumably unexpected or surprising) events; (2) generate one or more hypothesis about these events; (3) judge the plausibility of the hypotheses; and (4) select the `best' hypothesis as the explanation \cite{hoffman2017explaining}.

Research in philosophy and cognitive science has argued that abductive reasoning is closely related to explanation. In particular, in trying to understand causes of events, people use abductive inference to determine what they consider to be the ``best'' explanation. \citeA{harman1965inference} is perhaps the first to acknowledge this link, and more recently, experimental evaluations have demonstrated it \cite{lombrozo2012explanation,wilkenfeld2015inference,lombrozo2014explanation,rehder2003causal}. \citeA{popper2005logic} is perhaps the most influential proponent of abductive reasoning in the scientific process. He argued strongly for the scientific method to be based on empirical falsifiability of hypotheses, rather than the classic inductivist view at the time.

Early philosophical work considered abduction as some magical process of intuition --- something that could not be captured by formalised rules because it did not fit the standard deductive model. However, this changed when artificial intelligence researchers began investigating abductive reasoning to explain observations, such as in diagnosis (e.g. medical diagnosis, fault diagnosis) \cite{pople1973mechanization,reiter1987theory}, intention/plan recognition \cite{charniak1991probabilistic}, etc.
 The necessity to encode the process in a suitable computational form led to axiomatisations, with \citeA{pople1973mechanization} seeming to be the first to do this, and characterisations of how to implement such axiomatisations; e.g. \citeA{levesque1989knowledge}. From here, the process of abduction as a principled process gained traction, and it is now widely accepted that abduction, induction, and deduction are different modes of logical reasoning.

In this paper, abductive inference is not equated directly to explanation, because explanation also refers to the product and the social process; but abductive reasoning does fall into the category of cognitive process of explanation.  In Section~\ref{sec:cognitive-processes}, we survey the cognitive science view of abductive reasoning, in particular, cognitive biases in hypothesis formation and evaluation.
\end{changed}

%

\subsubsection{Interpretability and Justification}

Here, we briefly address the distinction between \emph{interpretability}, \emph{explainability}, \emph{justification}, and \emph{explanation}, as used in this article; and as they seem to be used in artificial intelligence. 

\citeA{lipton2016mythos} provides a taxonomy of the desiderata and methods for interpretable AI. This paper adopts \citeauthor{lipton2016mythos}'s assertion that explanation is post-hoc interpretability. I use \citeA{biran2017explanation}'s definition of \emph{interpretability} of a model as: the degree to which an observer can understand the cause of a decision. Explanation is thus one mode in which an observer may obtain understanding, but clearly, there are additional modes that one can adopt, such as making decisions that are inherently easier to understand or via introspection. I equate `interpretability' with `explainability'.

A \emph{justification} explains why a decision is good, but does not necessarily aim to give an explanation of the actual decision-making process \cite{biran2017explanation}.

It is important to understand the similarities and differences between these terms as one reads this article, because some related research discussed is  relevant to explanation only, in particular, Section~\ref{sec:social-explanation}, which discusses how people present explanations to one another; while other sections, in particular Sections~\ref{sec:social-attribution} and \ref{sec:cognitive-processes}  discuss how people generate and evaluate explanations, and explain behaviour of others, so are broader and can be used to create more explainable agents.

\subsection{Why People Ask for Explanations}
\label{sec:philosophical-foundations:why-ask-for-explanations}

There are many reasons that people may ask for explanations. Curiosity is one primary criterion that humans use, but other pragmatic reasons include examination --- for example, a teacher asking her students for an explanation on an exam for the purposes of testing the students' knowledge on a particular topic; and scientific explanation --- asking why we observe  a particular environmental phenomenon. 

In this paper, we are interested in explanation in AI, and thus our focus is on how intelligent agents can explain their decisions. As such, this section is primarily concerned with why people ask for `everyday' explanations of why \emph{specific} events occur, rather than explanations for general scientific phenomena, although this work is still relevant in many cases.

It is clear that the primary function of explanation is to facilitate \emph{learning} \cite{lombrozo2006structure,williams2013hazards}. Via learning, we obtain better models of how particular events or properties come about, and we are able to use these models to our advantage. \citeA{heider1958psychology} states that people look for explanations to improve their understanding of someone or something so that they can derive stable model that can be used for prediction and control.  This hypothesis is backed up by research suggesting that people tend to ask questions about events or observations that they consider abnormal or unexpected from their own point of view \cite{hilton1986knowledge,hilton1996mental,hesslow1988problem}.

\citeA{lombrozo2006structure} argues that explanations have a role in inference learning precisely \emph{because} they are explanations, not necessarily just due to the causal information they reveal. First, explanations provide somewhat of a `filter' on the causal beliefs of an event. Second, prior knowledge is changed by giving explanations; that is, by asking someone to provide an explanation as to whether a particular property is true or false, the explainer changes their perceived likelihood of the claim. Third, explanations that offer fewer causes and explanations that explain multiple observations are considered more believable and more valuable; but this does not hold for causal statements. \citeA{wilkenfeld2015inference} go further and show that engaging in explanation but failing to arrive at a correct explanation can improve ones understanding. They describe this as ``\emph{explaining for the best inference}'', as opposed to the typical model of explanation as ``\emph{inference to the best explanation}''.

\citeA[Chapter~3]{malle2004mind}, who gives perhaps the most complete discussion of everyday explanations in the context of explaining social action/interaction,  argues that people ask for explanations for two reasons:

\begin{enumerate}
\item \emph{To find meaning}: to reconcile the contradictions or inconsistencies between elements of our knowledge structures.
\item \emph{To manage social interaction}: to create a \emph{shared meaning} of something, and to change others' beliefs \& impressions, their emotions, or to influence their actions.
\end{enumerate}

Creating a shared meaning is important for explanation in AI. In many cases, an explanation provided by an intelligent agent will be precisely to do this --- to create a shared understanding of the decision that was made between itself and a human observer, at least to some partial level.

\citeA{lombrozo2006structure} and  \citeA{wilkenfeld2015inference} note that explanations have several functions other than the transfer of knowledge, such as persuasion, learning, or assignment of blame; and that in some cases of social explanation, the goals of the explainer and explainee may be different. With respect to explanation in AI, persuasion is surely of interest: if the goal of an explanation from an intelligent agent is to generate trust from a human observer, then persuasion that a decision is the correct one could in some case be considered more important than actually transferring the true cause. For example, it may be better to give a less likely explanation that is more convincing to the explainee if we want them to act in some positive way. In this case, the goals of the explainer (to generate trust) is different to that of the explainee (to understand a decision).

\subsection{Contrastive Explanation}
\label{sec:philosophical-foundations:contrastive-explanation}

\begin{quote}
The key insight is to recognise that one does not explain events per se, but that one explains why the puzzling event occurred in the target cases but not in some counterfactual contrast case. --- \citeA[p.\ 67]{hilton1990conversational}
\end{quote}

I will dedicate a subsection to discuss one of the most important findings in the philosophical and cognitive science literature from the perspective of explainable AI: \emph{contrastive explanation}. Research shows that people do not explain the causes for an event \emph{per se}, but explain the cause of an event \emph{relative to some other event} that did not occur; that is, an explanation is always of the form ``\emph{Why \P rather than \Q?}'', in which \P is the target event and \Q is a counterfactual contrast case that did not occur, even if the \Q is implicit in the question. This is called \emph{contrastive explanation}.

Some authors refer to \Q as the \emph{counterfactual case} \cite{lombrozo2012explanation,hesslow1988problem,hilton1986knowledge}. However, it is important to note that this is not the same counterfactual that one refers to when determining causality (see Section~\ref{sec:philosophical-foundations:causality}). For causality, the counterfactuals are hypothetical `non-causes' in which the event-to-be-explained does not occur --- that is a counterfactual to cause $C$ ---, whereas in contrastive explanation, the counterfactuals are hypothetical outcomes --- that is, a counterfactual to event $E$ \cite{mcgill1993contrastive}.

\citeA{lipton1990contrastive} refers to the two cases, \P and \Q, as the \emph{fact} and the \emph{foil} respectively; the \emph{fact} being the event that did occur, and the foil being the event that did not. To avoid confusion, throughout the remainder of this paper, we will adopt this terminology and use \emph{counterfactual} to refer to the hypothetical case in which the cause \C did not occur, and \emph{foil} to refer to the hypothesised case \Q that was expected rather than \P.

Most authors in this area argue that \emph{all} why--questions ask for contrastive explanations, even if the foils are not made explicit \cite{lipton1990contrastive,hilton1986knowledge,hesslow1988problem,hilton1990conversational,mackie1980cement,lombrozo2012explanation}, and that people are good at inferring the foil; e.g.\ from language and tone. For example, given the question, ``\emph{Why did Elizabeth open the door?}'', there are many, possibly an infinite number, of foils; e.g.\ ``\emph{Why did Elizabeth open the door, \emph{rather than} leave it closed?}'', ``\emph{Why did Elizabeth open the door \emph{rather than} the window?''}, or ``\emph{Why did Elizabeth open the door \emph{rather than} Michael opening it?}''. These different contrasts have different explanations, and there is no inherent one that is certain to be the foil for this question. The negated presupposition \emph{not(Elizabeth opens the door)} refers to an entire class of foils, including all those listed already. \citeA{lipton1990contrastive} notes that ``central requirement for a sensible contrastive question is that the fact and the foil have a largely similar history, against which the differences stand out. When the histories are disparate, we do not know where to begin to answer the question.'' This implies that people could use the similarity of the history of facts and possible foils to determine what the explainee's foil truly is.

It is important that the explainee understands the counterfactual case \cite{hesslow1988problem}. For example, given the question ``\emph{Why did Elizabeth open the door?}'', the answer ``\emph{Because she was hot}'' is a good answer if the foil is Elizabeth leaving the door closed, but not a good answer if the foil is ``\emph{\emph{rather than} turning on the air conditioning}'', because the fact that Elizabeth is hot explains both the fact and the foil. 

The idea of contrastive explanation should not be controversial if we accept the argument outlined in Section~\ref{sec:philosophical-foundations:why-ask-for-explanations} that people ask for explanations about events or observations that they consider abnormal or unexpected from their own point of view \cite{hilton1986knowledge,hilton1996mental,hesslow1988problem}. In such cases, people expect to observe a particular event, but then observe another, with the observed event being the fact and the expected event being the foil.

\begin{changed}
\citeA{van2002remote} define four types of explanatory question, three of which are contrastive:

{\centering

\vspace{2mm}

\noindent
\begin{tabular}{lp{11.5cm}}
  \emph{Plain fact}: &  Why does object $a$ have property $P$?\\
  \emph{P-contrast}: & Why does object $a$ have property $P$, rather than property $Q$?\\
  \emph{O-contrast}: & Why does object $a$ have property $P$, while object $b$ has property $Q$?\\
  \emph{T-contrast}: &  Why does object a have property $P$ at time $t$, but property $Q$ at time $t'$?
\end{tabular}
}

\citeauthor{van2002remote} note that differences occur on properties within an  object (P-contrast), between objects themselves (O-contrast), and within an object over time (T-contrast). They reject the idea that all `plain fact' questions have an implicit foil, proposing that plain-fact questions require showing details across a `non-interrupted' causal chain across time. They argue that plain-fact questions are typically asked due to curiosity, such as desiring to know how certain facts fit into the world, while contrastive questions are typically asked when unexpected events are observed.
\end{changed}

\citeA{lipton1990contrastive} argues that contrastive explanations between a fact \P and a foil \Q are, in general, easier to derive than `complete' explanations for plain-fact questions about \P. For example, consider the arthropod classification algorithm in Section~\ref{sec:intro:example}. To be a beetle, an arthropod must have six legs, but this does not cause an arthropod to be a beetle -- other causes are necessary. \citeauthor{lipton1990contrastive} contends that we could answer the P-contrast question such as ``\emph{Why is image \emph{J} labelled as a Beetle instead of a \emph{Spider}}?'' by citing the fact that the arthropod in the image has six legs. We do not need information about eyes, wings, or stingers to answer this, whereas to explain why image \emph{J} is a spider in a non-contrastive way, we must cite all causes.



The hypothesis that all causal explanations are contrastive is not merely philosophical. In Section~\ref{sec:cognitive-processes}, we see several bodies of work supporting this, and these provide more detail as to how people select and evaluate explanations based on the contrast between fact and foil.

\subsection{Types and Levels of Explanation}
\label{sec:philosophical-foundations:levels}

The type of explanation provided to a question is dependent on the particular question asked; for example, asking why some event occurred is different to asking under what circumstances it \emph{could have} occurred; that is, the actual vs.\ the hypothetical \cite{salmon2006four}. However, for the purposes of answering why--questions, we will focus on a particular subset of philosophical work in this area.

Aristotle's \emph{Four Causes} model, also known as the \emph{Modes of Explanation} model, continues to be foundational for cause and explanation. Aristotle proposed an analytic scheme, classed into four different elements, that can be used to provide answers to why--questions \cite{hankinson2001cause}:

\begin{enumerate}
\item \emph{Material}: The substance or material of which something is made. For example, rubber is a material cause for a car tyre. 
\item \emph{Formal}: The form or properties of something that make it what it is. For example, being round is a formal cause of a car tyre. These are sometimes referred to as \emph{categorical} explanations.
\item \emph{Efficient}: The proximal mechanisms of the cause something to change. For example, a tyre manufacturer is an efficient cause for a car tyre. These are sometimes referred to as \emph{mechanistic} explanations.
\item \emph{Final}: The end or goal of something.  Moving a vehicle is an efficient cause of a car tyre. These are sometimes referred to as \emph{functional} or \emph{teleological} explanations.
\end{enumerate}
 
A single why--question can have explanations from any of these categories. For example, consider the question: ``\emph{Why does this pen contain ink?}''. A material explanation is based on the idea that the pen is made of a substance that prevents the ink from leaking out. A formal explanation is that it is a pen and pens contain ink. An efficient explanation is that there was a person who filled it with ink. A final explanation is that pens are for writing, and so require ink.

\begin{short}
Several other authors have proposed models similar to Aristotle's, such as \citeA{dennett1989intentional}, who proposed that people take three \emph{stances} towards objects: \emph{physical}, \emph{design}, and \emph{intention}; and \citeA{marr1982vision}, building on earlier work with Poggio \cite{marr1976understanding}, who define the \emph{computational}, \emph{representational}, and \emph{hardware} levels of understanding for computational problems.

\citeA{kass1987types} define a categorisation of explanations of anomalies into three types: (1) \emph{intentional}; (2) \emph{material}; and (3) \emph{social}. The intentional and material categories correspond roughly to Aristotle's final and material categories, however, the \emph{social} category does not correspond to any particular category in the models of Aristotle, \citeA{marr1982vision}, or \citeA{dennett1989intentional}. The social category refers to explanations about human behaviour that is not intentionally driven. \citeauthor{kass1987types} give the example of an increase in crime rate in a city, which, while due to intentional behaviour of individuals in that city, is not a phenomenon that can be said to be intentional. While individual crimes are committed with intent, it cannot be said that the individuals had the intent of increasing the crime rate --- that is merely an effect of the behaviour of a group of individuals.
\end{short}

\begin{extended}

\citeA{dennett1989intentional} identifies three \emph{stances} that people seem to take towards an object:

\begin{enumerate}
  \item \emph{Physical}: The structure and form of the object.
  \item \emph{Design}: The mechanisms used for an object to carry out its function.
 \item \emph{Intentional}: The `Theory of Mind' that people hold about an object, attributing intentions, beliefs, and desires to it.
\end{enumerate}

These three categories relate closely to elements in Aristotle's model, except that Dennet's \emph{Intentional} stance implies, unlike Aristotle's model, that objects themselves take on intentions, rather than just their designers.

With regards to understanding of computational processes, in particular computational vision, \citeA{marr1982vision}, building on earlier work with Poggio \cite{marr1976understanding},  argues that computational processes need to be understood at three\footnote{\protect The original model contained four categories due to  \citeA{marr1976understanding} split \emph{Representation and Algorithm} into separate categories.} different levels:

\begin{enumerate}
 \item \emph{Computational theory}: This level describes \emph{what} a process does and \emph{why}. The operation that fulfils the `what' is defined in terms of a set of constraints that it must satisfy, but importantly, does not describe how to calculate this.
 \item \emph{Representation and algorithm}: This level describes the \emph{representations} of the input and output space of process, and an \emph{algorithm} for calculating the output from the input such that it satisfies the constraints in the high level. A given computational theory may have many representations and algorithms, but the choice of the representation and algorithm are closely linked to each other.
 \item \emph{Hardware implementation}: This level describes how the representation and algorithm are physically realised in a specific architecture.
\end{enumerate}

Again, there is quite a clear link with Aristotle's Four Causes model.  

\citeA{marr1982vision} contends that while these levels are related --- for example, the choice of an algorithm is clearly dependent on the computational specification that it has to solve, and is influenced by the hardware on which it must run --- the levels are only loosely coupled, in that almost any theory has many possible representations and algorithms, and such algorithms work on many types of hardware. Further, \citeauthor{marr1982vision} argues that \emph{explanation}s at each level are largely independent of the others, and some phenomenon may be described by appealing to just one level. Despite this, \citeauthor{marr1982vision} is clear that we need to understand the interactions between these levels, and study all three levels simultaneously.

\citeA{poggio2012levels} argues that his earlier model with \citeauthor{marr1982vision} \cite{marr1976understanding,marr1982vision} was a suitable frame of mind for computation science at that time, but that given advances in the field, two additional levels may be required. Thus, he proposes the inclusion of two additional levels on top of Marr's original levels of analysis:

\begin{enumerate}
 \item \emph{Learning} (on top of computational theory): This level describes approaches that are able to learn how to solve problems on their own. The ability to learn a solution from some data frees one from having to know the nuances of the computational aspects of solving the problem.
 \item \emph{Evolution} (on top of learning): This level describes learning approaches that can themselves evolve, thus knowing how to evolve learning machines frees one from having to understand the nuances of how learning approaches work.
\end{enumerate} 

\citeA{poggio2012levels} places learning on top of computational theory, in that if we learn how to solve a problem, we need not understand the computational definition. 
However, learning approaches devise new algorithms for solving a given computational problem --- they do not derive new computational problems to solve. On the other hand, evolution, allows us to solve \emph{new} computational problems, not only new learning problems. In effect, the evolution is not about evolving learning, but about evolving which computational problems should be solved.

At the end of his article, \citeA{poggio2012levels} adds two comments from (presumably) readers: one who states that learning has computational, algorithmic, and mechanistic properties, and thus learning is not a layer but just an instance of the previous model. Figure~\ref{fig:learning-in-marrs-levels}, taken from  \citeA{poggio2012levels}, outlines this. The second comment argues that evolution and learning are in fact the same thing, and thus should be combined. We return to this discussion in Section~\ref{sec:philosophical-foundations:AI} when we discuss the link of this work to artificial intelligence.

\begin{figure}[!ht]
\centering
\tikzstyle{centerednode} = [text width=10em, text centered]
\tikzstyle{leftnode} = [text width=4em]

\begin{tikzpicture}
   \begin{scope}[node distance=2cm and 5cm,on grid]
   \node [centerednode] (corticallearning)     {cortical learning\\ mechanisms};
   \node [centerednode] (corticalvision)	     [right=of corticallearning]  {cortical filtering\\mechanisms} edge [<->] (corticallearning);
   \node [leftnode] (mechanism) [left=of corticallearning] {Mechanism};

   \node [centerednode] (particularlearning)  [above=of corticallearning]   {particular learning\\ algorithms} edge [<->] (corticallearning);
   \node [centerednode] (channelbased)	     [above=of corticalvision]  {e.g.\ channel-based\\matching algorithms} edge [<->] (corticalvision)  edge [<->] (particularlearning);
   \node [leftnode] (algorithm) [above=of mechanism] {Algorithm};

   \node [centerednode] (mathematicsoflearning)  [above=of particularlearning]   {mathematics of\\learning}  edge [<->] (particularlearning);
   \node [centerednode] (stereo)	     [above=of channelbased]  {e.g.\ stereo or shape\\recognition theory} edge  [<->] (channelbased) edge [<->] (mathematicsoflearning);
   \node [leftnode] (computationaltheory) [above=of algorithm] {Computational\\Theory};

  \node [centerednode] (learning)  [above=of mathematicsoflearning]   {\textbf{~\\Learning}};
   \node [centerednode] (visionmodel)	     [above=of stereo]  {\textbf{~\\Vision Module}};

   \end{scope}
\end{tikzpicture}
\caption{Learning as a specific instance rather than a level. Taken from \citeA{poggio2012levels} who attributes it to Keith Nishihara.}
\label{fig:learning-in-marrs-levels}
\end{figure}

\citeA{kass1987types} define a categorisation of explanations of anomalies into three types: (1) \emph{intentional}; (2) \emph{material}; and (3) \emph{social}. The intentional and material categories correspond roughly for Aristotle's final and material categories, however, the \emph{social} category does not correspond to any particular category in the models of Aristotle, \citeA{marr1982vision}, or \citeA{dennett1989intentional}. The social category refers to explanations about human behaviour that is not intentionally driven. \citeauthor{kass1987types} give the example of an increase in crime rate in a city, which, while due to intentional behaviour of individuals in that city, is not a phenomenon that can be said to be intentional. While individual crimes are committed with intent, it cannot be said that the individuals had the intent of increasing the crime rate --- that is merely an effect of the behaviour of a group of individuals.

\citeA{overton2011scientific} presents an insightful model into the types of explanation. In particular, he defines three types of presuppositions for why--questions:

\begin{enumerate}
 \item Particular instances of processes, events, or objects;
 \item \emph{Phenomena}; that is patterns over different processes, events, of objects.
 \item \emph{Models} of phenomena.
\end{enumerate}

The first type can be considered presuppositions for `everyday' explanations, while the latter two are more about scientific explanation.  Different levels of explanation can be given for the different types of presuppositions. \citeA{overton2011scientific} proposes four different types of explanation, based on the different \emph{type signatures} of the presuppositions and answers:

\begin{enumerate}
 \item \emph{Design/causation}: The history of a particular design causes the property of a particular processes, event, or object.
 \item \emph{Instantiation}: A phenomenon is instantiated and the instantiation has the properties of the phenomenon.
 \item \emph{Modelling}: A model produces a certain phenomenon.
 \item \emph{Argumentation/justification}: Laws, definitions, approximations produce a model.
\end{enumerate}

\citeA{overton2011scientific} argues that these \emph{kinds} of explanations are different to \emph{levels} of explanation.
\end{extended}

\subsection{Structure of Explanation}
\label{sec:philosophical-foundations:structure}

As we saw in Section~\ref{sec:philosophical:definitions:explanation}, causation is a major part of explanation. Earlier accounts of explanation from \citeA{hempel1948studies} argued for logically deductive models of explanation. \citeA{kelley1967attribution} subsequently argued instead that people consider \emph{co-variation} in constructing explanations, and proposed a \emph{statistical} model of explanation. However, while influential, subsequent experimental research uncovered many problems with these models, and currently, both the deductive and statistical models of explanation are no longer considered valid theories of everyday explanation in most camps \cite{malle2011time}.

\begin{figure}[!th]
\centering

\begin{tikzpicture}[>=stealth, thick, scale=0.95]
\node (t) at (0,0) {Theories};
\node (m) at (3,0) {Models};
\node (k) at (6,0) {Kinds};
\node (e) at (9,0) {Entities};
\node (d) at (12,0) {Data};

\node(tm) at (1.5,1) {\it justifies};
       \draw[->] (t) to[out=70,in=180] (tm.south) to[out=0,in=110] (m);
\node(mk) at (4.5,1) {\it models};
       \draw[->] (m) to[out=70,in=180] (mk.south) to[out=0,in=110] (k);
\node(ke) at (7.5,1) {\it instantiated by};
       \draw[->] (k) to[out=70,in=180] (ke.south) to[out=0,in=110] (e);
\node(de) at (10.5,1) {\it measured by};
       \draw[->] (e) to[out=70,in=180] (de.south) to[out=0,in=110] (d);

\node (t2) at (0,-2)  {\it unifies};
       \draw[->] (t) to[out=290,in=0] (t2.north) to[out=180,in=250] (t);
\node (m2) at (3,-2) {\it submodel of};
       \draw[->] (m) to[out=290,in=0] (m2.north) to[out=180,in=250] (m);
\node (k2) at (6,-2)   {\it subkind of};
       \draw[->] (k) to[out=290,in=0] (k2.north) to[out=180,in=250] (k);
\node (e2) at (9,-2) {\it causes};
       \draw[->] (e) to[out=290,in=0] (e2.north) to[out=180,in=250] (e);
\node (d2) at (12,-2)   {\it correlates with};
       \draw[->] (d) to[out=290,in=0] (d2.north) to[out=180,in=250] (d);
\end{tikzpicture}
\caption{Overton's five categories and four relations in scientific explanation, reproduced from \citeA[p.\ 54, Figure 3.1]{overton2012explanation}}
\label{fig:philosophical-foundations:overton-categories}.
\end{figure}
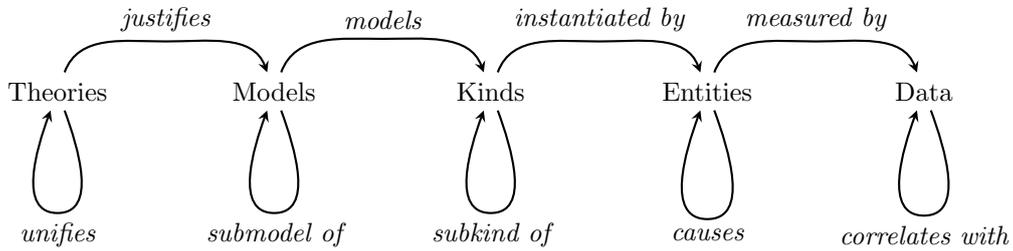

\citeA{overton2013explain,overton2012explanation} defines a model of scientific explanation. In particular, \citeA{overton2012explanation} defines the \emph{structure} of explanations. He defines five categories of properties or objects that are explained in science: (1) \emph{theories}: sets of principles that form building blocks for models; (2) \emph{models}: an abstraction of a theory that represents the relationships between kinds and their attributes; (3) \emph{kinds}: an abstract universal class that supports counterfactual reasoning; (4) \emph{entities}: an instantiation of a kind; and (5) \emph{data}: statements about activities (e.g.\ measurements, observations). The relationships between these is shown in Figure~\ref{fig:philosophical-foundations:overton-categories}.

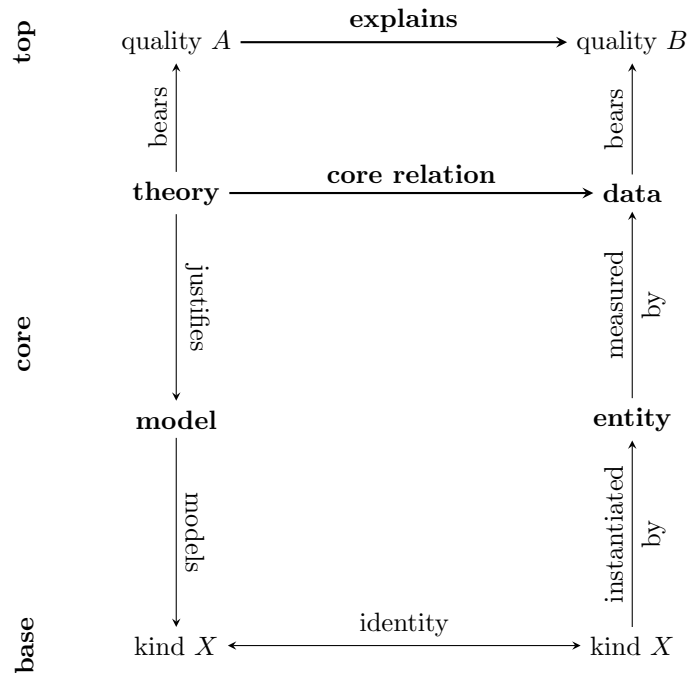
\begin{figure}[!th]
\centering

\begin{tikzpicture}[>=stealth]
\node [rotate=90] (a) at (0,8) {\bf top};
\node [rotate=90] (b) at (0,4) {\bf core};
\node [rotate=90] (c) at (0,0) {\bf base};

\node (1) at (2,8) {quality $A$};
\node (2) at (2,6) {\bf theory};
\node (3) at (2,3) {\bf model};
\node (4) at (2,0) {kind $X$};
\node (5) at (8,0) {kind $X$};
\node (6) at (8,3) {\bf entity};
\node (7) at (8,6) {\bf data};
\node (8) at (8,8) {quality $B$};

\draw [->] (2) -- (1) node[midway,above,sloped]{bears};
\draw [->] (7) -- (8) node[midway,above,sloped]{bears};

\draw [->] (2) -- (3) node[midway,above,sloped]{justifies};
\draw [->] (3) -- (4) node[midway,above,sloped]{models};
\draw [<->] (4) -- (5) node[midway,above,sloped]{identity};
\draw [->] (5) -- (6) node[midway,above,sloped]{instantiated}
		       node[midway,below,sloped]{by};
\draw [->] (6) -- (7) node[midway,above,sloped]{measured}
		       node[midway,below,sloped]{by};

\draw [->,thick] (1) -- (8) node[midway,above,sloped]{\bf explains};
\draw [->,thick] (2) -- (7) node[midway,above,sloped]{\bf core relation};
\end{tikzpicture}
\caption{Overton's general structure of a theory-data explanation, reproduced from \citeA[p.\ 54, Figure 3.2]{overton2012explanation})}
\label{fig:philosophical-foundations:overton-structure}
\end{figure}

From these categories, \citeA{overton2012explanation} provides a crisp definition of the structure of scientific explanations. He argues that explanations of phenomena at one level must be relative to and refer to at least one other level, and that explanations between two such levels must refer to all intermediate levels. For example, an arthropod (\emph{Entity}) has eight legs (\emph{Data}). Entities of this \emph{Kind} are spiders, according to the \emph{Model} of our \emph{Theory} of arthropods. In this example, the explanation is constructed by appealing to the \emph{Model} of insects, which, in turn, appeals to a particular \emph{Theory} that underlies that \emph{Model}.   Figure~\ref{fig:philosophical-foundations:overton-structure} shows the structure of a \emph{theory-data explanation}, which is the most complex because it has the longest chain of relationships between any two levels.

With respect to social explanation, \citeA{malle2004mind} argues that social explanation is best understood as consisting of three layers: 
\begin{enumerate}

 \item Layer 1: A conceptual framework that outlines the assumptions people make about human behaviour and explanation.

 \item Layer 2: The psychological processes that are used to construct explanations.

 \item Layer 3: Language layer that specifies the type of linguistic structures people use in giving explanations. 

\end{enumerate}

I will present \citeauthor{malle2004mind}'s views of these three layers in more detail in the section on social attribution (Section~\ref{sec:social-attribution}), cognitive processes (Section~\ref{sec:cognitive-processes}, and social explanation (Section~\ref{sec:social-explanation}). This work is collated into \citeauthor{malle2004mind}'s 2004 book \cite{malle2004mind}.

\subsection{Explanation and XAI}
\label{sec:philosophical-foundations:AI}

This section presents some ideas on how the philosophical work outlined above affects researchers and practitioners in XAI.

\subsubsection{Causal Attribution is Not Causal Explanation}

An important concept is the relationship between cause attribution and explanation. Extracting a causal chain and displaying it to a person is causal attribution, not (necessarily) an explanation. While a person could use such a causal chain to obtain their own explanation, I argue that this does not constitute giving an explanation. In particular, for most AI models, it is not reasonable to expect a lay-user to be able to interpret a causal chain, no matter how it is presented. Much of the existing work in explainable AI literature is on the causal attribution part of explanation --- something that, in many cases, is the easiest part of the problem because the causes are well understood, formalised, and accessible by the underlying models. In later sections, we will see more on the difference between attribution and explanation, why existing work in causal attribution is only part of the problem of explanation, and insights of how this work can be extended to produce more intuitive explanations.

\subsubsection{Contrastive Explanation}
\label{sec:philosophical-foundations:AI:contrastive-explanation}

Perhaps the most important point in this entire section is that explanation is contrastive (Section~\ref{sec:philosophical-foundations:contrastive-explanation}). Research indicates that people request only contrastive explanations, and that the cognitive burden of complete explanations is too great.

It could be argued that because models in AI operate at a level of abstraction that is considerably higher than real-world events, the causal chains are often smaller and less cognitively demanding, especially if they can be visualised. Even if one agrees with this, this argument misses a key point: it is not only the size of the causal chain that is important --- people seem to be cognitively wired to process contrastive explanations, so one can argue that a layperson will find contrastive explanations more intuitive and more valuable.


\begin{changed}
This is both a challenge and an opportunity in AI.
It is a challenge because often a person may just ask ``Why X?'', leaving their foil implicit. Eliciting a contrast case from a human observer may be difficult or even infeasible.  \citeA{lipton1990contrastive} states that the obvious solution is that a non-contrastive question ``\emph{Why \P?}''  can be interpreted by default to ``\emph{Why \P rather than \notP?''}. However, he then goes on to show that to answer ``\emph{Why \P rather than \notP?''} is equivalent to providing all causes for \P --- something that is not so useful. As such, the challenge is that the foil needs to be determined. In some applications, the foil could be elicited from the human observer, however, in others, this may not be possible, and therefore, foils may have to be inferred.  As noted later in Section~\ref{sec:cognitive-processes:AI:abnormality}, concepts such as \emph{abnormality} could be used to infer likely foils, but techniques for HCI, such as eye gaze \cite{singh2018combining} and gestures could be used to infer foils in some applications.

It is an opportunity because, as \citeA{lipton1990contrastive} argues, explaining a contrastive question is often easier than giving a full causal attribution because one only needs to understand what is \emph{different} between the two cases, so one can provide a complete explanation without determining or even knowing all of the causes of the fact in question. This holds for computational explanation as well as human explanation.

Further, it can be beneficial in a more pragmatic way: if a person provides a foil, they are implicitly pointing towards the part of the model they do not understand. 
In Section~\ref{sec:cognitive-processes:explanation-selection}, we will see research that outlines how people use contrasts to select explanations that are much simpler than their full counterparts.

Several authors within artificial intelligence flag the importance of contrastive questions.
\citeA{lim2009assessing} found via a series of user studies on context-aware applications that ``\emph{Why not \ldots?}'' questions were common questions that people asked. Further, several authors have looked to answer contrastive questions. For example,
\citeA{winikoff2017debugging} considers the questions of ``\emph{Why don't you believe \ldots?}'' and ``\emph{Why didn't you do \ldots?}'' for BDI programs,  or \citeA{fox2017explainable} who have similar questions in planning, such as ``\emph{Why didn't you do something else (that I would have done)}?''. However, most existing work considers contrastive \emph{questions}, but not contrastive \emph{explanations}; that is, finding the differences between the two cases. Providing two complete explanations does not take advantage of contrastive questions. Section~\ref{sec:cognitive-processes:explanation-selection:facts-and-foils} shows that people use the difference between the fact and foil to \emph{focus} explanations on the causes relevant to the question, which makes the explanations more \emph{relevant} to the explainee.
\end{changed}

\subsubsection{Explanatory Tasks and Levels of Explanation}
\label{sec:philosophical-foundations:discussions:levels}

Researchers and practitioners in explainable AI should understand and adopt a model of `levels of explanation' --- either one of those outlined above, or some other sensible model. The reason is clear: the answer that is provided to the why--question is strongly linked to the level at which the question is posed.

To illustrate, let's take a couple of examples and apply them to Aristotle's modes of explanation model outlined in Section~\ref{sec:philosophical-foundations:levels}. Consider our earlier arthropod classification algorithm from Section~\ref{sec:intro:example}.  At first glance, it may seem that such an algorithm resides at the \emph{formal} level, so should offer explanations based on form. However, this would be erroneous, because that given categorisation algorithm has both efficient/mechanistic components, a reason for being implemented/executed (the \emph{final} mode), and is implemented on hardware (the \emph{final} mode). As such, there are explanations for its behaviour at all levels.  
Perhaps most why--questions proposed by human observers about such an algorithm would indeed by at the formal level, such as ``\emph{Why is image J in group A instead of group B?}'', for which an answer could refer to the particular form of image and the groups A and B. However, in our idealised dialogue, the question ``\emph{Why did you infer that the insect in image J had eight legs instead of six?}'' asks a question about the underlying algorithm for counting legs, so the cause is at the efficient level; that is, it does not ask for what constitutes a spider in our model, but from where the inputs for that model came. Further, the final question about classifying the spider as an octopus refers to the final level, referring to the algorithms \emph{function} or \emph{goal}. Thus, causes in this algorithm occur at all four layers: (1) the material causes are at the hardware level to derive certain calculations; (2) the formal causes determine the classification itself; (3) the efficient causes determine such concepts as how features are detected; and (4) final causes determine why the algorithm was executed, or perhaps implemented at all.

As a second example, consider an algorithm for planning a robotic search and rescue mission after a disaster. In planning, programs are dynamically constructed, so different modes of cause/explanation are of interest compared to a classification algorithm. Causes still occur at the four levels: (1) the material level as before describes the hardware computation; (2) the formal level describes the underlying model passed to the planning tool; (3) the mechanistic level describes the particular planning algorithm employed; and (4) the final level describes the particular goal or intention of a plan. In such a system, the robot would likely have several goals to achieve; e.g.\ searching, taking pictures, supplying first-aid packages, returning to re-fuel, etc. As such, why--questions described at the final level (e.g.\ its goals) may be more common than in the classification algorithm example. However, questions related to the model are relevant, or why particular actions were taken rather than others, which may depend on the particular optimisation criteria used (e.g.\ cost vs.\ time), and these require efficient/mechanistic explanations.

 However, I am not arguing that we, as practitioners, must have explanatory agents capable of giving explanations at all of these levels. I argue that these frameworks are useful for analysing the types of questions explanatory agents one may receive. In Sections~\ref{sec:social-attribution} and \ref{sec:cognitive-processes}, we will see work that demonstrates that for explanations at these different levels, people expect different types of explanation. Thus, it is important to understand which types of questions refer to which levels in particular instances of technology, that different levels will be more useful/likely than others, and that, in research articles on interpretability, it is clear at which level we are aiming to provide explanations.


\subsubsection{Explanatory Model of Self}
\label{sec:philosophical-foundations:model-of-self}

The work outlined in this section demonstrates that an intelligent agent must be able to reason about its own causal model. Consider our image classification example. When posed with the question ``\emph{Why is image J in group A instead of group B?}'', it is non-trivial, in my view, to attribute the cause by using the algorithm that generated the answer. A cleaner solution would be to have a more abstract symbolic model alongside this that records information such as when certain properties are detected and when certain categorisations are made, which can be reasoned over. In other words, the agent requires a model of it's own decision making --- a \emph{model of self} --- that exists merely for the purpose of explanation. This model may be only an approximation of the original model,  but more suitable for explanation.

\begin{changed}
This idea is not new in XAI. In particular, researchers have investigated machine learning models that are uninterpretable, such as neural nets, and have attempted to extract model approximations using more interpretable model types, such as Bayesian networks \cite{harradon2018causal}, decision trees \cite{frosst2017distilling}, or local approximations \cite{ribeiro2016should}.  However, my argument here is not only for the purpose of interpretability. Even models considered interpretable, such as decision trees, could be accompanied by another model that is specifically used for explanation. For example, to explain control policies, \citeA{hayes2017improving} select and annotate particular important state variables and actions that are relevant \emph{for explanation only}. \citeauthor{langley2017explainable} notes that ``\emph{An agent must represent content in a way that supports the explanations}'' \cite[p.\ 2]{langley2017explainable}.
\end{changed}

Thus, to generate meaningful and useful explanations of behaviour, models based on the our understanding of explanation must sit alongside and work with the decision-making mechanisms.

\subsubsection{Structure of Explanation}
\label{sec:philosophical-foundations:structure-of-explanation}

\begin{changed}
Related to the `model of self' is the structure of explanation. \citeauthor{overton2012explanation}'s model of scientific explanation  \cite{overton2012explanation} defines what I believe to be a solid foundation for the structure of explanation in AI. To provide an explanation along the chain outlined in Figure~\ref{fig:philosophical-foundations:overton-structure}, one would need an explicit explanatory model (Section~\ref{sec:philosophical-foundations:model-of-self}) of \emph{each of these different categories} for the given system.


For example, the question from our dialogue in Section~\ref{sec:intro:example} ``\emph{How do you know that spiders have eight legs?}'', is a question referring not to the causal attribution in the classification algorithm itself, but is asking: ``\emph{How do you know this?}'', and thus is referring to how this was learnt --- which, in this example, was learnt via another algorithm.  Such an approach requires an additional part of the `model of self' that refers specifically to the learning, not the classification.
\end{changed}



\citeauthor{overton2012explanation}'s model \cite{overton2012explanation} or one similar to it seems necessary for researchers and practitioners in explainable AI to frame their thoughts and communicate their ideas.


\section{Social Attribution --- How Do People Explain Behaviour?}
\label{sec:social-attribution}

\begin{quote}
Just as the contents of the nonsocial environment are interrelated by certain lawful connections, causal or otherwise, which define what can or will happen, we assume that there are connections of similar character between the contents of the social environment. -- \citeA[Chapter 2, pg.\ 21]{heider1958psychology}
\end{quote}

In this section, we outline work on \emph{social attribution}, which defines how people attribute and (partly) explain behaviour of others. Such work is clearly relevant in many areas of artificial intelligence. However, research on social attribution laid the groundwork for much of the work outlined in Section~\ref{sec:cognitive-processes}, which looks at how people generate and evaluate events more generally. For a more detailed survey on this, see \citeA{mcclure2002goal} and \citeA{hilton2017social}.

\subsection{Definitions}

Social attribution is about \emph{perception}. While the causes of behaviour can be described at a neurophysical level, and perhaps even lower levels, social attribution is concerned not with the real causes of human behaviour, but how other attribute or explain the behaviour of others. \citeA{heider1958psychology} defines social attribution as \emph{person perception}.

\emph{Intentions} and \emph{intentionality} is key to the work of \citeA{heider1958psychology}, and much of the recent work that has followed his --- for example, \citeA{dennett1989intentional,malle2004mind,mcclure2002goal,boonzaier2005distinguishing,kashima1998category}. An intention is a mental state of a person in which they form a commitment to carrying out some particular action or achieving some particular aim.  \citeA{malle1997folk} note that intentional behaviour therefore is always contrasted with \emph{unintentional} behaviour, citing that laws of state, rules in sport, etc.\ all treat intentional actions different from unintentional actions because intentional rule breaking is punished more harshly than unintentional rule breaking. They note that, while intentionality can be considered an \emph{objective} fact, it is also a \emph{social} construct, in that people ascribe intentions to each other whether that intention is objective or not, and use these to socially interact.

\emph{Folk psychology}, or commonsense psychology, is the attribution of human behaviour using `everyday' terms such as beliefs, desires, intentions, emotions, and personality traits. This field of cognitive and social psychology recognises that, while such concepts may not truly cause human behaviour, these are the concepts that humans use to model and predict each others' behaviours \cite{malle2004mind}. In other words, folk psychology does not describe how we think; it describes how we think we think.

In the folk psychological model, actions consist of three parts: (1) the precondition of the action --- that is, the circumstances under which it can be successfully executed, such as the capabilities of the actor or the constraints in the environment; (2) the action itself that can be undertaken; and (3) the effects of the action --- that is, the changes that they bring about, either environmentally or socially.

Actions that are undertaken are typically explained by \emph{goals} or intentions. In much of the work in social science, \emph{goals} are equated with intentions. For our discussions, we define \emph{goals} as being the end to which  a mean contributes, while we define \emph{intentions} as short-term goals that are adopted to achieve the end goals. The intentions have no utility themselves except to achieve positive utility goals.  A \emph{proximal} intention is a near-term intention that helps to achieve some further \emph{distal} intention or goal. In the survey of existing literature, we will use the term used by the original authors, to ensure that they are interpreted as the authors expected.

\subsection{Intentionality and Explanation}
\label{sec:social-attribution:intentionality}

\begin{figure}[!th]
\centering
\includegraphics[scale=0.4]{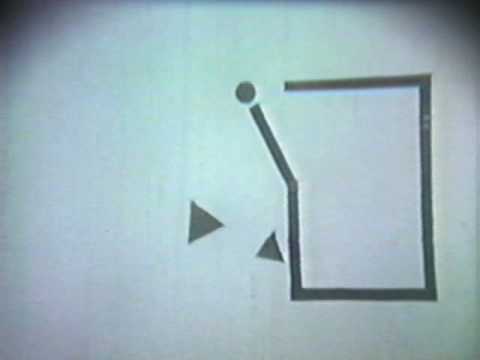}
\caption{A screenshot of the video used in \citeauthor{heider1944anexperimental}'s seminal study \cite{heider1944anexperimental}.}
\label{fig:social-attribution:heider-simmel-screenshot}
\end{figure}

\citeA{heider1958psychology} was the first person to experimentally try to identify how people attribute behaviour to others. In their now famous experiment from 1944, \citeA{heider1944anexperimental}, showed a video containing animated shapes --- a small triangle, a large triangle, and a small circle --- moving around a screen\footnote{See the video here: \url{https://www.youtube.com/watch?v=VTNmLt7QX8E}.}, and asked experiment participants to observe the video and then  describe the behaviour of the shapes. Figure~\ref{fig:social-attribution:heider-simmel-screenshot} shows a captured screenshot from this video in which the circle is opening a door to enter into a room. The participants' responses described the behaviour anthropomorphically, assigning actions, intentions, emotions, and personality traits to the shapes. However, this experiment was not one on animation, but in social psychology. The aim of the experiment was to demonstrate that people characterise deliberative behaviour using folk psychology.

\citeA{heider1958psychology} argued then that, the difference between \emph{object perception} --- describing causal behaviour of objects --- and person perception was the intentions, or \emph{motives}, of the person. He noted that behaviour in a social situation can have two types of causes: (1) \emph{personal} (or \emph{dispositional}) causality; and (2) \emph{impersonal} causality, which can subsequently be influenced by \emph{situational} factors, such as the environment.  This interpretation lead to many researchers reflecting on the \emph{person-situation} distinction and, in \citeauthor{malle2011time}'s view \cite{malle2011time}, incorrectly interpreting \citeauthor{heider1958psychology}'s work for decades. 
\begin{extended}
\citeauthor{malle2011time} claims that the prevailing view (at the time of his research in the late 1990s and early 2000s) of the person-situation distinction of explanation is misplaced due to people misinterpreting \citeauthor{heider1958psychology}'s work, and instead argues for a distinction based around \emph{intentional vs.\ unintentional} behaviour. While \citeauthor{heider1958psychology} did indeed make a distinction between person and situation in the cause of events, he seemed more interested in the distinction between intentional vs.\ unintentional, which he called \emph{personal and impersonal causality}.
\end{extended}

\citeA{heider1958psychology} contends that the key distinction between intentional action and non-intentional events is that intentional action demonstrates \emph{equifinality}, which states that while the means to realise an intention may vary, the intention itself remains equa-final. Thus, if an actor should fail to achieve their intention, they will try other ways to achieve this intention, which differs from physical causality. \citeA{lombrozo2010causal} provides the example of Romeo and Juliet, noting that had a wall been placed between them, Romeo would have scaled the wall or knocked in down to reach his goal of seeing Juliet. However, iron filaments trying to  get to a magnet would not display such equifinality --- they would instead be simply blocked by the wall. Subsequent research confirms this distinction \cite{dennett1989intentional,malle2004mind,mcclure2002goal,boonzaier2005distinguishing,kashima1998category,lombrozo2012explanation}.

\begin{extended}
\citeA[Chapter~1]{malle2004mind} argues that Heider's view was that the distinction between personal vs.\ situational attribution is used to explain how action \emph{outcomes} were attained, but not \emph{why} a person is acting in a particular way. That is, given a particular outcome of an observed action, one refers to the personal capabilities of an agent to achieve a certain outcome (internal or personal), as well as to the environmental factors that may have contributed to that outcome (external or situational).
\end{extended}

\citeA{malle2001attention}  break the actions that people will explain into two dimensions: (1) \emph{intentional vs.\ unintentional}; and (2) \emph{observable vs.\ unobservable}; thus creating four different classifications (see Figure~\ref{fig:malles-classificiation-of-types-of-events}).

\begin{figure}[!h]
\centering
\begin{tabular}{lll}
\toprule
   & \textbf{Intentional} & \textbf{Unintentional}\\
\midrule
\textbf{Observable}   & actions & mere behaviours \\
\textbf{Unobservable} & intentional thoughts & experiences \\
\bottomrule
\end{tabular}
\caption{Malle's classification of types of events, based on the dimensions of intentionality and observability \cite[Chapter~3]{malle2004mind}}
\label{fig:malles-classificiation-of-types-of-events}
\end{figure}

\citeA{malle2001attention} performed experiments to confirm this model. As part of these experiments, participants were placed into a room with another participant, and were left for 10 minutes to converse with each other to `get to know one another', while their conversation was recorded. \citeauthor{malle2001attention} coded participants responses to questions with regards to observability and intentionality. Their results show that actors tend to explain unobservable events more than observable events, which \citeauthor{malle2001attention} argue is because the actors are more \emph{aware} of their own beliefs, desires, feelings, etc.,\ than of their observable behaviours, such as facial expressions, gestures, postures, etc.). On the other hand, observers do the opposite for the inverse reason. Further, they showed that actors tend to explain unintentional behaviour more than intentional behaviour, again because (they believe) they are aware of their intentions, but not their `unplanned' unintentional behaviour. Observers tend to find both intentional and unintentional behaviour difficult to explain, but will tend to find intentional behaviour more \emph{relevant}. Such a model accounts for the \emph{correspondence bias} noted by \citeA{gilbert1995correspondence}, which is the tendency for people to explain others' behaviours based on traits rather than situational factors, because the situational factors (beliefs, desires) are invisible.

\subsection{Beliefs, Desires, Intentions, and Traits}

Further to intentions, research suggest that other factors are important in attribution of social behaviour; in particular, beliefs, desires, and traits.

\citeA{kashima1998category} demonstrated that people use the folk psychological notions of belief, desire, and intention to understand, predict, and explain human action. In particular, they demonstrated that desires hold preference over beliefs, with beliefs being not explained if they are clear from the viewpoint of the explainee. They showed that people judge that explanations and behaviour `do not make sense' when belief, desires, and intentions were inconsistent with each other. This early piece of work is one of the first to re-establish \citeauthor{heider1958psychology}'s theory of intentional behaviour in attribution \cite{heider1958psychology}.

However, it is the extensive body of work from \citeauthor{malle2004mind} \cite{malle1999people,malle2004mind,malle2011attribution} that is the most seminal in this space. 

\subsubsection{Malle's Conceptual Model for Social Attribution}

 \citeA{malle2004mind} proposes a model based on \emph{Theory of Mind}, arguing that people attribute behaviour of others and themselves by assigning particular mental states that explain the behaviour. He offers six postulates (and sub-postulates) for the foundation of people's folk explanation of behaviour, modelled in the scheme in Figure~\ref{fig:malles-conceptual-framework}. He argues that these six postulates represent the assumptions and distinctions that people make when attributing behaviour to themselves and others:

\begin{figure}[!h]
\centering
\includegraphics[scale=0.5]{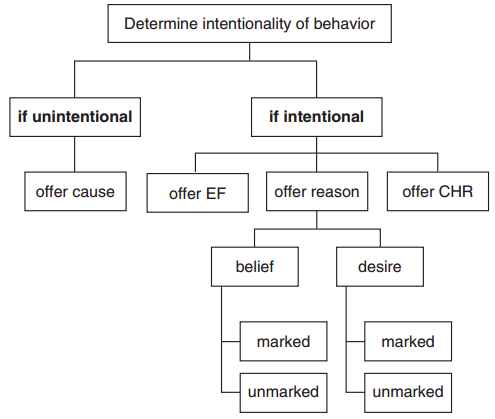}
\caption{Malle's conceptual framework for behaviour explanation; reproduced \citeA[p.\ 87, Figure 3.3]{malle2011attribution}, adapted from \citeA[p.\ 119, Figure 5.1]{malle2004mind}}
\label{fig:malles-conceptual-framework}
\end{figure}

\begin{enumerate}

 \item People distinguish between intentional and unintentional behaviour.

 \item For intentional behaviour, people use three modes of explanation based on the specific circumstances of the action:

   \begin{enumerate}

    \item \emph{Reason explanations} are those explanations that link to the mental states (typically desires and beliefs, but also values) for the act, and the grounds on which they formed an intention.
     
    \item \emph{Causal History of Reason (CHR) explanations} are those explanations that use factors that ``lay in the background'' of an agent's reasons (note, not the background of the action), but are not themselves reasons. Such factors can include unconscious motives, emotions, culture, personality, and the context. CHR explanations refer to causal factors that lead to reasons.

  CHR explanations do not presuppose either subjectivity or rationality. This has three implications. First, they do not require the explainer to take the perspective of the explainee. Second, they can portray the actor as less rationale, by not offering a rational and intentional reason for the behaviour. Third, they allow the use of unconscious motives that the actor themselves would typically not use. Thus, \emph{CHR explanations can make the agent look less rationale and in control than reason explanations}.

    \item \emph{Enabling factor (EF) explanations} are those explanations that explain not the intention of the actor, but instead explain how the intentional action achieved the outcome that it did. Thus, it assumes that the agent had an intention, and then refers to the factors that enabled the agent to successfully carry out the action, such as personal abilities or environmental properties. In essence, it relates to why preconditions of actions were enabled.
   \end{enumerate}

 \item For unintentional behaviour, people offer just \emph{causes}, such as physical, mechanistic, or habitual cases.

\end{enumerate}

At the core of \citeauthor{malle2004mind}'s framework is the intentionality of an act. For a behaviour to be considered intentional, the behaviour must be based on some \emph{desire}, and a belief that the behaviour can be undertaken (both from a personal and situational perspective) and can achieve the desire. This forms the \emph{intention}. If the agent has the ability and the awareness that they are performing the action, then the action is intentional.


Linguistically, people make a distinction between causes and reasons; for example, consider ``\emph{What were her reasons for choosing that book?}'', vs.\ ``\emph{What were his causes for falling over?}''.  The use of ``\emph{his causes}'' implies that the cause does not belong to the actor, but the reason does.

To give a reason explanation is to attribute \emph{intentionality} to the action, and to identify the desires, beliefs, and valuings \emph{in light of which} (subjectivity assumption) and \emph{on the grounds of which} (rationality assumption) the agent acted. Thus, reasons imply intentionality, subjectivity, and rationality.

\subsection{Individual vs.\ Group Behaviour}
\label{sec:social-attribution:groups}

\citeA{susskind1999perceiving} investigated how people ascribe causes to groups rather than individuals, focusing on traits. They provided experimental participants with a set of statements describing behaviours performed by individuals or groups, and were then asked to provide ratings of different descriptions of these individuals/groups, such as their intelligence (a trait, or CHR in Malle's framework), and were asked to judge the confidence of their judgements.  Their results showed that as with individuals, participants freely assigned traits to groups, showing that groups are seen as agents themselves. However, they showed that when explaining an individual's behaviour, the participants were  able to produce explanations faster and more confidently than for groups, and that the traits that they assigned to individuals were judged to be less `extreme' than those assigned to to groups. In a second set of experiments, \citeauthor{susskind1999perceiving} showed that people expect more consistency in an individual's behaviour compared to that of a group. When presented with a behaviour that \emph{violated} the impression that participants had formed of individuals or groups, the participants were more likely to attribute the individual's behaviour to causal mechanisms than the groups' behaviour.

\citeA{olaughlin2002people} further investigated people's perception of group vs.\ individual behaviour, focusing on intentionality of explanation. They investigated the relative agency of groups that consist of `unrelated' individuals acting independently (\emph{aggregate groups}) compared to groups acting together (\emph{jointly acting groups}). 
In their study, participants were more likely to offer CHR explanations than intention explanations for aggregate groups, and more likely to offer intention explanations than CHR explanations for \emph{jointly} acting groups. For instance, to explain why all people in a department store came to that particular store, participants were more likely offer a CHR explanation, such as that there was a sale on at the store that day. However, to answer the same question for why a group of friends came to the same store place, participants were more likely to offer an explanation that the group wanted to spend the day together shopping -- a desire. This may demonstrate that people cannot attribute intentional behaviour to the individuals in an aggregate group, so resort to more causal history explanations.

\citeauthor{olaughlin2002people}'s \cite{olaughlin2002people} finding about using CHRs to explain aggregate group behaviour is consistent with the earlier work from \citeA{kass1987types}, whose model of explanation explicitly divided \emph{intentional} explanations from \emph{social explanations}, which are explanations about human behaviour that is not intentionally driven (discussed in more detail in Section~\ref{sec:philosophical-foundations:levels}). These social explanations account for how people attribute deliberative behaviour to groups without referring to any form of intention.

An intriguing result from \citeA{olaughlin2002people} is that while people attribute less intentionality to aggregate groups than to individuals, they attribute \emph{more} intentionality to jointly acting groups than to individuals. \citeauthor{olaughlin2002people} reason that joint action is highly deliberative, so the group intention is more likely to have been explicitly agreed upon prior to acting, and the individuals within the group would be explicitly aware of this intention compared to the their own individual intentions.

\subsection{Norms and Morals}
\label{sec:social-attribution:norms}

\emph{Norms} have been shown to hold a particular place in social attribution. \citeA{burguet2004effets} (via \citeA{hilton2017social}) showed that norms and abnormal behaviour are important in how people ascribe mental states to one another. For example, \citeA{hilton2017social} notes that upon hearing the statement ``\emph{Ted admires Paul}'', people tend to attribute some trait to Paul as the object of the sentence, such as that Paul is charming and many people would admire him; and even that Ted does not admire many people. However, a counter-normative statement such as ``\emph{Ted admires the rapist}'' triggers attributions instead to Ted, explained by the fact that it is non-normative to admire rapists, so Ted's behaviour is distinctive to others, and is more likely to require an explanation. In Section~\ref{sec:cognitive-processes}, we will see more on the relationship between norms, abnormal behaviour, and attribution.

\citeA{uttich2010norms} investigate the relationship of norms and the effect it has on attributing particular mental states, especially with regard to morals. They offer an interesting explanation of \emph{the side-effect effect}, or the \emph{Knobe effect} \cite{knobe2003intentional}, which is the effect for people to attribute particular mental states (Theory of Mind) based on moral judgement. Knobe's vignette from his seminal \cite{knobe2003intentional} paper is: 

\begin{quote}
The vice-president of a company went to the chairman of the board and said, ``We are thinking of starting a new program. It will help us increase profits, but it will also harm the environment''. The chairman of the board answered, ``I don't care at all about harming the environment. I just want to make as much profit as I can. Let's start the new program.'' They started the new program. Sure enough, the environment was harmed.
\end{quote}

Knobe then produce a second vignette, which is exactly the same, but the side-effect of the program was in fact that the environment was \emph{helped}. When participants were asked if the chairman had \emph{intentionally} harmed the environment (first vignette), 82\% of respondents replied yes. However, in the second vignette, only 23\% thought that the chairman intentionally helped the environment.

\citeA{uttich2010norms} hypothesis that the two existing camps aiming to explain this effect: the \emph{Intuitive Moralist} and \emph{Biased Scientist}, do not account for this.  \citeauthor{uttich2010norms} hypothesise that it is the fact the \emph{norms} are violated that account for this; that is, rather than moralist judgements influencing intentionality attribution, it is the more general relationship of conforming (or not) to norms (moral or not). In particular, behaviour that conforms to norms is less likely to change a person's Theory of Mind (intention) of another person compared to behaviour that violates norms. 

\citeA{samland2014social} further investigate social attribution in the context of norms, looking at permissibility rather than obligation. They gave participants scenarios in which two actors combined to cause an outcome. For example, a department in which only administrative assistants are permitted to take pens from the stationary cupboard. One morning, Professor Smith (not permitted) and an assistant (permitted) each take a pen, and there are no pens remaining. Participants were tasked with rating how strongly each agent caused the outcome. Their results showed that participants rated the action of the non-permitted actor (e.g.\ Professor Smith) more than three times stronger than the other actor. However, if the outcome was positive instead of negative, such as an intern (not permitted) and a doctor (permitted) both signing off on a request for a drug for a patient, who subsequently recovers due to the double dose, participants rate the non-permitted behaviour only slightly stronger. As noted by \citeA[p.\ 54]{hilton2017social}, these results indicate that in such settings, people seem to interpret the term \emph{cause} as meaning ``morally or institutionally responsible''.

In a follow-up study, \citeA{samland2016role} showed that children are not sensitive to  norm violating behaviour in the same way that adults are. In particular, while both adults and children correlate cause and blame, children do not distinguish between cases in which the person was aware of the norm, while adults do.

\subsection{Social Attribution and XAI}
\label{sec:social-attribution:AI}

This section presents some ideas on how the work on social attribution outlined above affects researchers and practitioners in XAI.

\subsubsection{Folk Psychology}
\label{sec:social-attribution:AI:folk-psychology}

While the models and research results presented in this section pertain  to the behaviour of humans, it is reasonably clear that these models have a place in explainable AI.  \citeauthor{heider1944anexperimental}'s seminal  experiments from 1944 with moving shapes \cite{heider1944anexperimental} (Section~\ref{sec:social-attribution:intentionality})  demonstrate unequivocally that people attribute folk psychological concepts such as belief, desire, and intention, to artificial objects. Thus, as argued by \citeA{degraaf2017people}, it is not a stretch to assert that people will expect explanations using the same conceptual framework used to explain human behaviours.

This model is particularly promising because many knowledge-based models in deliberative AI either explicitly build on such folk psychological concepts, such as \emph{belief-desire-intention} (BDI) models \cite{rao1995bdi}, or can be mapped quite easily to them; e.g.\ in classical-like AI planning, goals represent desires, intermediate/landmark states represent intentions, and the environment model represents beliefs \cite{ghallab2004automated}.

In addition, the concepts and relationships between actions, preconditions, and proximal and distal intentions is similar to those in models such as BDI and planning, and as such, the work on the relationships between preconditions, outcomes, and competing goals, is useful in this area.

\subsubsection{Malle's Models}

Of all of the work outlined in this section, it is clear that \citeauthor{malle2004mind}'s model, culminating in his 2004 text book \cite{malle2004mind}, is the most mature and complete model of social attribution to date. His three-layer models provides a solid foundation on which to build explanations of many deliberative systems, in particular, goal-based deliberation systems.

\citeauthor{malle2004mind}'s conceptual framework provides a suitable framework for characterising different aspects of causes for behaviour. It is clear that reason explanations will be useful for goal-based reasoners, as discussed in the case of BDI models and goal-directed AI planning, and  enabling factor explanations can play a role in \emph{how} questions and in counterfactual explanations.  In Section~\ref{sec:cognitive-processes}, we will see further work on how to \emph{select} explanations based on these concepts. 

However, the causal history of reasons (CHR) explanations also have a part to play for deliberative agents. In human behaviour, they refer to personality traits and other unconscious motives. While anthropomorphic agents could clearly use CHRs to explain behaviour, such as emotion or personality, they are also valid explanations for non-anthropomorphic agents. For example, for AI planning agents that optimise some metric, such as cost, the explanation that action $a$ was chosen over action $b$ because it had lower cost is a CHR explanation. The fact that the agent is optimising cost is a `personality trait' of the agent that is invariant given the particular plan or goal. Other types of planning systems may instead be risk averse, optimising to minimise risk or regret, or may be `flexible' and try to help out their human collaborators as much as possible. These types of explanations are CHRs; even if they are not described as personality traits to the explainee. However, one must be careful to ensure these CHRs do not make their agent appear irrational --- unless of course, that is the goal one is trying to achieve with the explanation process.

\begin{changed}
\citeA{broekens2010you} describe algorithms for automatic generation of explanations for BDI agents. Although their work does not build on \citeauthor{malle2004mind}'s model directly, it shares a similar structure, as noted by the authors, in that their model uses intentions and enabling conditions as explanations. They present three algorithms for explaining behaviour: (a) offering the goal towards which the action contributes; (b) offering the enabling condition of an action; and (c) offering the next action that is to be performed; thus, the explanadum is explained by offering a proximal intention. A set of human behavioural experiments showed that the different explanations are considered better in different circumstances; for example, if only one action is required to achieve the goal, then offering the goal as the explanation is more suitable than offering the other two types of explanation, while if it is part of a longer sequence, also offering a proximal intention is evaluated as being a more valuable explanation. These results reflect those by \citeauthor{malle2004mind}, but also other results from social and cognitive psychology on the link between goals, proximal intentions, and actions, which are surveyed in Section~\ref{sec:cognitive-processes:intentionality-and-functionality}
\end{changed}

\subsubsection{Collective Intelligence}

The research into behaviour attribution of groups (Section~\ref{sec:social-attribution:groups}) is important for those working in collective intelligence; areas such as in multi-agent planning \cite{brafman2008one}, computational social choice \cite{chevaleyre2007short}, or argumentation \cite{besnard2008elements}. Although this line of work appears to be much less explored than attributions of individual's behaviour, the findings from \citeA{kass1987types}, \citeauthor{susskind1999perceiving}, and in particular \citeA{olaughlin2002people} that people assign intentions and beliefs to jointly-acting groups, and reasons to aggregate groups, indicates that the large body of work on attribution of individual behaviour could serve as a solid foundation for explanation of collective behaviour.

\subsubsection{Norms and Morals}

The work on norms and morals discussed in Section~\ref{sec:social-attribution:norms} demonstrates that normative behaviour, in particular, violation of such behaviour, has a large impact on the ascription of a Theory of Mind to actors. Clearly, for anthropomorphic agents, this work is important, but as with CHRs, I argue here that it is important for more `traditional' AI as well. 

First, the link with morals is important for applications that elicit ethical or social concerns, such as defence, safety-critical applications, or judgements about people. Explanations or behaviour in general that violate norms may give the impression of `immoral machines' --- whatever that can mean --- and thus, such norms need to be explicitly considered as part of explanation and interpretability.

Second, as discussed in Section~\ref{sec:philosophical-foundations:why-ask-for-explanations}, people mostly ask for explanations of events that they find unusual or abnormal \cite{hilton1986knowledge,hilton1996mental,hesslow1988problem}, and violation of normative behaviour is one such abnormality \cite{hilton1996mental}. Thus, normative behaviour is important in interpretability --- a statement that would not surprise those researchers and practitioners of normative artificial intelligence.

In Section~\ref{sec:cognitive-processes}, we will see that norms and violation of normal/normative behaviour is also important in the cognitive processes of people asking for, constructing, and evaluating explanations, and its impact on interpretability.


\section{Cognitive Processes --- How Do People Select and Evaluate Explanations?}
\label{sec:cognitive-processes}

\begin{quote}
There are as many causes of \emph{x} as there are explanations of \emph{x}. Consider how the cause of death might have been set out by the physician as `multiple haemorrhage', by the barrister as `negligence on the part of the driver', by the carriage-builder as `a defect in the brakelock construction', by a civic planner as `the presence of tall shrubbery at that turning'. None is more true than any of the others, but the particular context of the question makes some explanations more relevant than others. -- \citeA[p.\ 54]{hanson1965patterns}
\end{quote}

\citeA{mill1973system} is one of the earliest investigations of cause and explanation, and he argued that we make use of `statistical' correlations to identify cause, which he called the \emph{Method of Difference}. He argued that causal connection and explanation selection are essentially arbitrary and the scientifically/philosophically it is ``wrong'' to select one explanation over another, but offered several cognitive biases that people seem to use, including things like unexpected conditions, precipitating causes, and variability. Such \emph{covariation} models ideas were dominant in causal attribution, in particular, the work of \citeA{kelley1967attribution}. However, many researchers noted that the covariation models failed to explain many observations; for  example,  people can identify causes between events from a single data point \cite{mcgill1993contrastive,hilton2010selecting}; and therefore, more recently, new theories have displaced them, while still acknowledging that the general idea that people using co-variations is valid.

In this section, we look at these theories, in particular, we survey three types of cognitive processes used in explanation: (1) \emph{causal connection}, which is the process people use to identify the causes of events; (2) \emph{explanation selection}, which is the process people use to select a small subset of the identified causes as \emph{the} explanation; and (3) \emph{explanation evaluation}, which is the processes that an explainee uses to evaluate the quality of an explanation. Most of this research shows that people have certain \emph{cognitive biases} that they apply to explanation generation, selection, and evaluation. 

\subsection{Causal Connection, Explanation Selection, and Evaluation}

\begin{figure}[!h]
\centering
\includegraphics[scale=0.5]{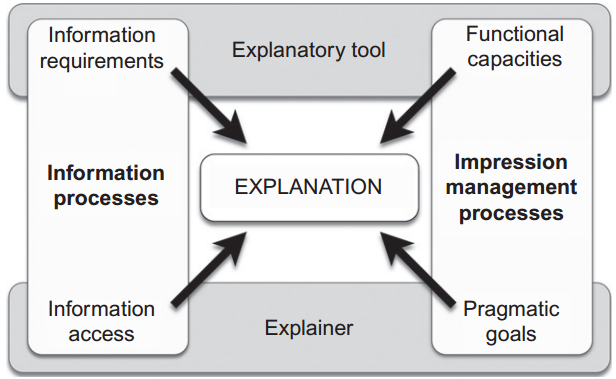}
\caption{Malle's process model for behaviour explanation; reproduced from \citeA[p.\ 320, Figure 6.6]{malle2011time}}
\label{fig:malles-process-model}
\end{figure}

\citeA{malle2004mind} presents a theory of explanation, which breaks the psychological processes used to offer explanations into two distinct groups, outlined in Figure~\ref{fig:malles-process-model}:

\begin{enumerate}

 \item \emph{Information processes} --- processes for devising and assembling explanations. The present section will present related work on this topic.

 \item \emph{Impression management processes} -- processes for governing the social interaction of explanation. Section~\ref{sec:social-explanation} will present related work on this topic.

\end{enumerate}

\citeA{malle2004mind} further splits these two dimensions into two further dimensions, which refer to the tools for constructing and giving explanations, and the explainer's perspective or knowledge about the explanation.

Taking the two dimensions, there are four items:

\begin{enumerate}
 \item \emph{Information requirements} --- what is required to give an adequate explanation; for example, one must knows the causes of the explanandum, such as the desires and beliefs of an actor, or the mechanistic laws for a physical cause.

 \item \emph{Information access} --- what information the explainer \emph{has} to give the explanation, such as the causes, the desires, etc. Such information can be lacking; for example, the explainer does not know the intentions or beliefs of an actor in order to explain their behaviour.

 \item \emph{Pragmatic goals} --- refers to the goal of the the explanation, such as transferring knowledge to the explainee, making an actor look irrational, or  generating trust with the explainee.

 \item \emph{Functional capacities} --- each explanatory tool has functional capacities that constrain or dictate what goals can be achieved with that tool.

\end{enumerate}

\citeA{malle2007actor} argue that this theory accounts for  apparent paradoxes observed in attribution theory, most specifically the actor-observer asymmetries, in which actors and observers offer different explanations for the same action taken by an actor. They hypothesise that  this is due to  \emph{information asymmetry}; e.g.\ an observer cannot access the intentions of an actor --- the intentions must be inferred from the actor's behaviour. 

In this section, we first look specifically at processes related to the explainer: information access and pragmatic goals. When requested for an explanation, people typically do not have direct access to the causes, but infer them from observations and prior knowledge. Then, they select some of those causes as the explanation, based on the goal of the explanation. These two process are known as \emph{causal connection} (or \emph{causal inference}), which is a processing of identifying the key causal connections to the fact; and \emph{explanation selection} (or \emph{casual selection}), which is the processing of selecting a subset of those causes to provide as an explanation.

\begin{changed}
This paper separates casual connection into two parts: (1) \emph{abductive reasoning}, the cognitive process in which people try to infer causes that explain events by making assumptions about hypotheses and testing these; and (2) \emph{simulation}, which is the cognitive process of simulating through counterfactuals to derive a good explanation. These processes overlap, but can be somewhat different. For example, the former requires the reasoner to make assumptions and test the validity of observations with respect to these assumptions, while in the latter, the reasoner could have complete knowledge of the causal rules and environment, but use simulation of counterfactual cases to derive an explanation. From the perspective of explainable AI, an explanatory agent explaining its decision would not require abductive reasoning as it is certain of the causes of its decisions. An explanatory agent trying to explain some observed events not under its control, such as the behaviour of another agent, may require abductive reasoning to find a plausible set of causes.
\end{changed}

Finally, when explainees \emph{receive} explanations, they go through the process of \emph{explanation evaluation}, through which they determine whether the explanation is satisfactory or not. A primary criteria is that the explanation allows the explainee to understand the cause, however, people's cognitive biases mean that they prefer certain types of explanation over others. 

\begin{changed}
\subsection{Causal Connection: Abductive Reasoning}

The relationship between explanation and abductive reasoning is introduced in Section~\ref{sec:philosophical-foundations:explanation:explanation-as-process}. This section surveys work in cognitive science that looks at the process of abduction. Of particular interest to XAI (and artificial intelligence in general) is work demonstrating the link between explanation and learning, but also other processes that people use to simplify the abductive reasoning process for explanation generation, and to switch modes of reasoning to correspond with types of explanation.
\end{changed}

\subsubsection{Abductive Reasoning and Causal Types}

\citeA{rehder2003causal} looked specifically at \emph{categorical} or \emph{formal} explanations. He presents the \emph{causal model theory}, which states that people infer categories of objects by both their features and the \emph{causal relationships between features}. His experiments show that people categorise objects based their perception that the observed properties were generated by the underlying causal mechanisms.  \citeauthor{rehder2003causal} gives the example that people not only know that birds can fly and bird have wings, but that birds can fly \emph{because} they have wings. 
In addition, \citeauthor{rehder2003causal} shows that people use combinations of features as evidence when assigning objects to categories, especially for features that seem incompatible based on the underlying causal mechanisms. For example, when categorising an animal that cannot fly, yet builds a nest in trees, most people would consider it implausible to categorise it as a bird because it is difficult to build a nest in a tree if one cannot fly. However, people are more likely to categorise an animal that does not fly and builds nests on the ground as a bird (e.g.\ an ostrich or emu), as this is more plausible; even though the first example has more features in common with a bird (building nests in trees).

\citeA{rehder2006similarity} extended this work to study how people \emph{generalise} properties based on the explanations received. When his participants were ask to infer their own explanations using abduction, they were more likely to generalise a property from a source object to a target object if they had more features that were similar; e.g. generalise a property from one species of bird to another, but not from a species of bird to a species of plant. However, given an explanation based on features, this relationship is almost completely eliminated: the generalisation was only done if the features detailed in the explanation were shared between the source and target objects; e.g. bird species $A$ and mammal $B$ both eat the same food, which is explained as the cause for an illness, for example. Thus, the abductive reasoning process used to infer explanations were also used to generalise properties -- a parallel seen in machine learning \cite{mitchell1986explanation}.

However, \citeA{williams2013hazards} demonstrate that, at least for categorisation in abductive reasoning, the properties of generalisation that support learning can in fact weaken learning by \emph{overgeneralising}. They gave experimental participants a categorisation task to perform by training themselves on exemplars. They asked one group to explain the categorisations as part of the training, and another to just `think aloud' about their task. The results showed that the explanation group more accurately categorised features that had similar patterns to the training examples, but less accurately categorised exceptional cases and those with unique features.  \citeauthor{williams2013hazards} argue that explaining (which forces people to think more systematically about the abduction process) is good for fostering generalisations, but this comes at a cost of over-generalisation.

\citeA{chin2017contrastive} provide support for the contrastive account of explanation (see Section~\ref{sec:philosophical-foundations:contrastive-explanation}) in categorisation/classification tasks. They hypothesise that \emph{contrast classes} (foils) are key to providing the context to explanation. They distinguish between \emph{prototypical} features of categorisation, which are those features that are typical of a particular category, and \emph{diagnostic} features, which are those features that are relevant for a contrastive explanation. Participants in their study were asked to either describe particular robots or explain why robots were of a particular category, and then follow-up on transfer learning tasks. The results demonstrated that participants in the design group mentioned significantly more features in general, while participants in the explanation group selectively targeted contrastive features. These results provide empirical support for contrastive explanation in category learning.

\subsubsection{Background and Discounting}

\citeA{hilton1996mental} discusses the complementary processes of \emph{backgrounding} and \emph{discounting} that affect the abductive reasoning process.
Discounting is when a hypothesis is deemed less likely as a cause because additional contextual information is added to a competing hypothesis as part of causal connection. It is actually discounted as a cause to the event.
Backgrounding involves pushing a possible cause to the background because it is not relevant to the goal, or new contextual information has been presented that make it no longer a good explanation (but still a cause). That is, while it is the cause of an event, it is not relevant to the explanation because e.g.\ the contrastive foil also has this cause.

As noted by \citeA{hilton1996mental}, discounting occurs in the context of multiple possible causes --- there are several possible causes and the person is trying to determine which causes the fact ---, while backgrounding occurs in the context of multiple necessary events --- a subset of necessary causes is selected as the explanation. Thus, discounting is part of causal connection, while backgrounding is part of explanation selection.

\subsubsection{Explanatory Modes}

As outlined in Section~\ref{sec:philosophical-foundations:levels}, philosophers and psychologists accept that different types of explanations exist; for example, Aristotle's model: material, formal, efficient, and final. However, theories of causality have typically argued for only one type of cause, with the two most prominent being dependence theories and transference theories. 

\citeA{lombrozo2010causal} argues that both dependence theories and transference theories are at least \emph{psychologically} real, even if only one (or neither) is the true theory. She hypothesises that people employ different modes of abductive reasoning for different modes of cognition, and thus both forms of explanation are valid: functional (final) explanations are better for phenomena that people consider have dependence relations, while mechanistic (efficient) explanations are better for physical phenomena.

\citeA{lombrozo2010causal} gave experimental participants scenarios in which the explanatory mode was manipulated and isolated using a mix of intentional and accidental/incidental human action, and in a second set of experiments, using biological traits that provide a particular function, or simply cause certain events incidentally. Participants were asked to evaluate different causal claims. The results of these experiments show that when events were interpreted in a functional manner, counterfactual dependence was important, but physical connections were not. However, when events were interpreted in a mechanistic manner, both counterfactual dependence and physical dependence were both deemed important. This implies that there is a link between functional causation and dependence theories on the one hand, and between mechanistic explanation and transference theories on the other.  The participants also rated the functional explanation stronger in the case that the causal dependence was intentional, as opposed to accidental.

\citeA{lombrozo2009explanation} studied at the same issue of functional vs.\ mechanistic explanations for inference in categorisation tasks specifically.  She presented participants with tasks similar to the following (text in square brackets added):

\begin{quote}
There is a kind of flower called a holing. Holings typically have brom compounds in their stems and they typically bend over as they grow. Scientists have discovered that having brom compounds in their stems is what usually causes holings to bend over as they grow \emph{[mechanistic cause]}. By bending over, the holing's pollen can brush against the fur of field mice, and spread to neighboring areas \emph{[functional cause]}.

\textbf{Explanation prompt}: \emph{Why do holings typically bend over?}
\end{quote}

They then gave participants a list of questions about flowers; for example: \emph{Suppose a flower has
brom compounds in its stem. How likely do you think it is that it bends over?}

Their results showed that participants who provided a mechanistic explanation from the first prompt were more likely to think that the flower would bend over, and vice-versa for functional causes. Their findings shows that giving explanations influences the inference process, changing the importance of different features in the understanding of category membership, and that the importance of features in explanations can impact the categorisation of that feature. In extending work, \citeA{lombrozo2014explanation} argue that people generalise better from functional than mechanistic explanations.

\begin{changed}
\subsubsection{Inherent and Extrinsic Features}
\label{sec:cognitive-processes:abductive-reasoninig:inherence}

\citeA{prasada2006principled} and \citeA{prasada2017scope} discuss how people's abductive reasoning process prioritises certain factors in the formal mode. \citeauthor{prasada2017scope} contends that ``\emph{Identifying something as an instance of a kind and explaining some of its properties in terms of its being the kind of thing it is are not two distinct activities, but a single cognitive activity.}'' \cite[p.\ 2]{prasada2017scope}

\citeA{prasada2006principled} note that people represent relationships between the kinds of things and the properties that they posses. This description conforms with \citeauthor{overton2012explanation}'s model of the structure of explanation \cite{overton2012explanation} (see Section~\ref{sec:philosophical-foundations:structure-of-explanation}).  \citeauthor{prasada2006principled}'s  experiments showed that people distinguish between two types of properties for a kind: \emph{k-properties}, which are the inherent properties of a thing that are due to its kind, and which they call \emph{principled connections}; and \emph{t-properties}, which are the extrinsic properties of a thing that are  not due to its kind, which they call \emph{factual connections}. Statistical correlations are examples of factual connections. For instance, a queen bee has a stinger and five legs because it is a bee (k-property), but the painted mark seen on almost all domesticated queen bees is because a bee keeper has marked it for ease of identification (t-property). K-properties have both principled and factual connections to their kind, whereas t-properties have mere factual connections. They note that k-properties have a \emph{normative} aspect, in that it is expected that instances of kinds will have their k-properties, and when they do not, they are considered abnormal; for instance, a bee without a stinger.

In their experiments, they presented participants with explanations using different combinations of k-properties and t-properties to explain categorisations; for example, ``why is this a dog?''
Their results showed that for formal modes, explanations involving k-properties were considered much better than explanations involving t-properties, and further, that using a thing's kind to explain why it has a particular property was considered better for explaining k-properties than for explaining t-properties. 

Using findings from previous studies, \citeA{cimpian2014inherence} argue that, when asked to explain a phenomenon, such as a feature of an object, people's cognitive biases make them more likely to use inherent features (k-properties) about the object to explain the phenomenon, rather than extrinsic features (t-properties), such as historical factors. An inherent feature is one that characterises ``how an object is constituted'' \cite[p.\ 465]{cimpian2014inherence}, and therefore they tend to be stable and enduring features. For example, ``spiders have eight legs'' is inherent, while ``his parents are scared of spiders'' is not. Asked to explain why they find spiders scary, people are more likely to refer to the ``legginess'' of spiders rather than the fact that their parents have arachnophobia, even though studies show that people with arachnophobia are more likely to have family members who find spiders scary \cite{davey1991characteristics}. \citeauthor{cimpian2014inherence} argue that, even if extrinsic information is known, it is not readily accessible by the \emph{mental shotgun} \cite{kahneman2011thinking} that people use to retrieve information. For example, looking at spiders, you can see their legs, but not your family's fear of them. Therefore, this leads to people biasing explanations towards inherent features rather than extrinsic. This is similar to the correspondence bias discussed in Section~\ref{sec:social-attribution:intentionality}, in which people are more likely to describe people's behaviour on personality traits rather than beliefs, desires, and intentions, because the latter are not readily accessible while the former are stable and enduring. The bias towards inherence is affected by many factors, such as prior knowledge, cognitive ability, expertise, culture, and age.
\end{changed}

\subsection{Causal Connection: Counterfactuals and Mutability}
\label{sec:cognitive-processes:mutability}

To determine the causes of anything other than a trivial event, it is not possible for a person to simulate back through all possible events and evaluate their counterfactual cases. Instead, people apply heuristics to select just some events to \emph{mutate}.  However, this process is not arbitrary. This section looks at several biases used to assess the \emph{mutability} of events; that is, the degree to which the event can be `undone' to consider counterfactual cases. It shows that abnormality (including social abnormality), intention, time and controllability of events are key criteria.

\subsubsection{Abnormality}

 \citeA{kahneman1982simulation} performed seminal work in this field, proposing the \emph{simulation heuristic}. They hypothesise that when answering questions about past events, people perform a mental simulation of counterfactual cases. In particular, they show that abnormal events are mutable: they are the common events that people undo when judging causality. In their experiments, they asked people to identity primary causes in causal chains using vignettes of a car accident causing the fatality of Mr.\ Jones, and which had multiple necessary causes, including Mr.\ Jones going through a yellow light, and the teenager driver of the truck that hit Mr.\ Jones' car while under the influence of drugs. They used two vignettes: one in which Mr.\ Jones the car took an unusual route home to enjoy the view along the beach (the \emph{route} version); and one in which he took the normal route home but left a bit early (the \emph{time} version). Participants were asked to complete an `if only' sentence that undid the fatal accident, imagining that they were a family member of Mr.\ Jones. Most participants in the route group  undid the event in which Mr.\ Jones took the unusual route home more than those in the time version, while those in the time version undid  the event of leaving early more often than those in the route version. That is, the participants tended to  focus more on \emph{abnormal} causes. In particular, \citeauthor{kahneman1982simulation} note that people did not simply undo the event with the lowest prior probability in the scenario.

In their second study, \citeA{kahneman1982simulation} asked the participants to empathise with the family of the teenager driving the truck instead of with Mr.\ Jones, they found that people more often undid events of the teenage driver, rather Mr.\ Jones. Thus, the \emph{perspective} or the \emph{focus} is important in what types of events people undo.


\subsubsection{Temporality}
\label{sec:cognitive-processes:temporality}

\citeA{miller1990temporal} show that the \emph{temporality} of events is important, in particular that people undo more recent events than more distal events. For instance, in one of their studies, they asked participants to play the role of a teacher selecting exam questions for a task. In one group, the \emph{teacher-first} group,  the participants were told that the students had not yet studied for their exam, while those in the another group, the \emph{teacher-second group}, were told that the students had already studied for the exam. Those in the teacher-second group selected easier questions than those in the first, showing that participants perceived the degree of blame they would be given for hard questions depends on the temporal order of the tasks.  This supports the hypothesis that earlier events are considered less mutable than later events.

\subsubsection{Controllability and Intent} 

\citeA{girotto1991event} investigated mutability in causal chains with respect to \emph{controllability}. They hypothesised that actions controllable by deliberative actors are more mutable than events that occur as a result of environmental effects. They provided participants with a vignette about Mr.\ Bianchi, who arrived late home from work to find his wife unconscious on the floor. His wife  subsequently died. Four different events caused Mr.\ Bianchi's lateness: his decision to stop at a bar for a drink on the way home, plus three non-intentional causes, such as delays caused by abnormal traffic. Different questionnaires were given out with the events in different orders. When asked to undo events, participants  overwhelmingly selected the intentional event as the one to undo, demonstrating that people mentally undo controllable events over uncontrollable events, irrelevant of the controllable events position in the sequence or whether the event was normal or abnormal. In another experiment, they varied whether the deliberative actions were \emph{constrained} or \emph{unconstrained}, in which an event is considered as constrained when they are somewhat enforced by other conditions; for example, Mr.\ Bianchi going to the bar (more controllable) vs.\ stopping due to an asthma attack (less controllable). The results of this experiment show that unconstrained actions are more mutable than constrained actions.

\subsubsection{Social Norms}

\citeA{mccloy2000counterfactual} investigated the mutability of controllable events further, looking at the perceived appropriateness (or the socially normative perception) of the events. They presented a vignette similar to that of \citeA{girotto1991event}, but with several controllable events, such as the main actor stopping to visit his parents, buy a newspaper, and stopping at a fast-food chain to get a burger. Participants were asked to provide causes as well as rate the `appropriateness' of the behaviour. The results showed that participants were more likely to indicate inappropriate events as causal; e.g.\ stopping to buy a burger. In a second similar study, they showed that inappropriate events are traced through both normal and other exceptional events when identifying cause.


\subsection{Explanation Selection}
\label{sec:cognitive-processes:explanation-selection}

Similar to causal connection, people do not typically provide all causes for an event as an explanation. Instead, they \emph{select} what they believe are the most relevant causes.
\citeA{hilton2017social} argues that explanation selection is used for cognitive reasons: causal chains are often too large to comprehend. He provides an example \cite[p.\ 43, Figure 7]{hilton2017social} showing the causal chain for the story of the fatal car accident involving `Mr.\ Jones' from \citeA{kahneman1982simulation}. For a simple story of a few paragraphs, the causal chain consists of over 20 events and 30 causes, all relevant to the accident. However, only a small amount of these are selected as explanations \cite{trabasso2003story}.

In this section, we overview key work that investigates the criteria people use for explanation selection. Perhaps unsurprisingly, the criteria for selection look similar to that of mutability, with temporality (proximal events preferred over distal events), abnormality, and intention being important, but also the features that are different between fact and foil.

\subsubsection{Facts and Foils}
\label{sec:cognitive-processes:explanation-selection:facts-and-foils}

As noted in Section~\ref{sec:philosophical-foundations}, why--questions are contrastive between a fact and a foil. Research shows that the two contrasts are the primary way that people \emph{select} explanations. In particular, to select an explanation from a set of causes, people look at the \emph{difference} between the cases of the fact and foil.

\citeA{mackie1980cement} is one of the earliest to argue for explanation selection based on contrastive criteria, however,  the first crisp definition of contrastive explanation seems to come from \citeA{hesslow1988problem}: 

\begin{quote}
This theory rests on two ideas. The first is that the effect or the explanandum, i.e.\ the event to be explained, should be construed, not as an object's having a certain property, but as a \emph{difference} between objects with regard to a certain property. The second idea is that selection and weighting of causes is determined by \emph{explanatory relevance}. [Emphasis from the original source] --- \citeA[p.\ 24]{hesslow1988problem}
\end{quote}

\citeA{hesslow1988problem} argues that criteria for selecting explanations are clearly not arbitrary, because people seem to select explanations in similar ways to each other. 
%
He defines an explanan as a relation containing an object $a$ (the \emph{fact} in our terminology), a set of comparison objects $R$, called the \emph{reference class} (the \emph{foils}), and a property $E$, which $a$ has but the objects in reference class $R$ does not. For example, $a =$ Spider, $R =$ Beetle, and $E= $ eight legs. 
 \citeauthor{hesslow1988problem} argues that the contrast between the fact and foil is the primary criteria for explanation selection, and that the explanation with the highest \emph{explanatory power} should be the one that highlights the greatest number of \emph{differences} in the attributes between the target and reference objects.

\begin{short}
\citeA{lipton1990contrastive}, building on earlier work in philosophy from \citeA{lewis1986causal}, derived similar thoughts to \citeA{hesslow1988problem}, without seeming to be aware of his work. He proposed a definition of contrastive explanation based on what he calls the \emph{Difference Condition}:

\begin{quote}
To explain why \P rather than \Q, we must cite a causal difference between \P and \notQ, consisting of a cause of \P and the absence of a corresponding event in the history of \notQ. -- \citeA[p.\ 256]{lipton1990contrastive}
\end{quote}

\end{short}

\begin{extended}
\citeA{lipton1990contrastive}, building on earlier work in philosophy from
\citeA{lewis1986causal}, derives similar thoughts to \citeA{hesslow1988problem}, without seeming to be aware of his work. \citeauthor{lewis1986causal} states that to explain why \P occurred rather than \Q, one should offer an event in the history of \P that would not have applied to the history of \Q, if \Q had occurred. For example, having eight legs is a cause of the image being categorised as a spider instead of a beetle because having eight legs is a feature of a spider, but would not be a feature of a beetle \emph{if} the outcome had been beetle.

However, \citeA{lipton1990contrastive} notes that this only works for \emph{incompatible} contrasts -- that is, when it is not possible that the fact and the foil could both happen. For cases in which the fact and foil are compatible, this definition does not work. For example, consider a similar case in which the fact is image \emph{J} being categorised as a fly, but the previous image, image \emph{K}, was categorised as a beetle. In this case, the why--question is asking why the current image was categorised as one type while the previous was another. However, in general, these could both be flies. The explanation that both have six legs is unsuitable, because it does not highlight a difference. However, \citeauthor{lipton1990contrastive} argues that under \citeauthor{lewis1986causal}'s definition \cite{lewis1986causal}, this would be a contrastive explanation: the event of \emph{J} having six legs is not a cause for categorising image \emph{K} as a beetle.

\citeA{lipton1990contrastive} proposes a definition of contrastive explanation based on what he calls the \emph{Difference Condition}:

\begin{quote}
To explain why \P rather than \Q, we must cite a causal difference between \P and \notQ, consisting of a cause of \P and the absence of a corresponding event in the history of \notQ. -- \citeA[p.\ 256]{lipton1990contrastive}
\end{quote}

This differs from \citeauthor{lewis1986causal}'s definition \cite{lewis1986causal} in that instead of selecting a cause of \P that is not a cause of \Q \emph{if} \Q had occurred, we should explain the \emph{actual difference} between \P and \notQ; that is, we should cite a cause that is in the actual history of \P, and contrast it with another event that did not occur in the actual history of \notQ.
We can formalise \citeauthor{lewis1986causal} and \citeauthor{lipton1990contrastive}'s definitions as the following, in which $\causes$ is the causal relation, and \HP and \HNOTQ are the \emph{history} of \P and \notQ respectively, and \HQ is the hypothetical history of \Q had it occurred:

\begin{center} 
\begin{tabular}{ll@{$~\land~$}l@{~~~where~~~}l@{$~\land~$}l}
\toprule
\citeauthor{lewis1986causal}       & $c \causes P$ & $c \not\causes Q$ & $c \in $~\HP & $c \notin $~\HQ\\
\citeauthor{lipton1990contrastive} & $c \causes P$ & $c' \causes $~\Q & $c \in $~\HP & $c' \notin $~\HNOTQ\\
\bottomrule
\end{tabular}
\end{center}

Thus, \citeauthor{lewis1986causal}'s definition \cite{lewis1986causal} cites some alternative history of facts in which \Q occurred, whereas \citeauthor{lipton1990contrastive}'s definition \cite{lipton1990contrastive} refers to the \emph{actual} history of \notQ. 
\end{extended}

From an experimental perspective, \citeA{hilton1986knowledge} were the first researchers to both identify the limitations of covariation, and instead propose that contrastive explanation is best described as the differences between the two events (discussed further in Section~\ref{sec:cognitive-processes:abnormality}). More recent research in cognitive science from \citeA{rehder2003causal,rehder2006similarity} supports the theory  that people perform causal inference, explanation, and generalisation based on contrastive cases.

Returning to our arthropod example, for the why--question between image \emph{J} categorised as a fly and image \emph{K} a beetle, image \emph{J} having six legs is correctly determined to have no explanatory relevance, because it does not cause \emph{K} to be categorised as a beetle instead of a fly. Instead, the explanation would cite some other cause, which according to Table~\ref{tab:arthropod}, would be that the arthropod in image \emph{J} has five eyes, consistent with a fly, while the one in image \emph{K} has two, consistent with a beetle.

\begin{extended}
\begin{figure}[!th]
\centering
\begin{minipage}{0.45\textwidth}
\centering
\begin{tikzpicture}
\begin{scope}[node distance=1.2cm and 0.03cm]

\tikzstyle{every initial by arrow}=[initial text=]
\tikzstyle{every state}=[circle,fill=none,draw=black,text=black,inner sep=0pt,minimum size=2.5mm]
\tikzstyle{white state}=[circle,fill=none,draw=black,text=black,inner sep=0pt,minimum size=2.5mm]
\tikzstyle{empty state}=[fill=none,inner sep=0pt,minimum size=2.5mm,opacity=0.0]
\tikzstyle{grey state}=[circle,fill=black,draw=black,text=black,inner sep=0pt,minimum size=2.5mm]

\node[white state] (T1) {};
\node[empty state,right=of T1] (T1g) {};
\node[grey state,right=of T1g] (T2) {};
\node[empty state,right=of T2] (T2g) {};
\node[white state,right=of T2g] (T3) {};
\node[empty state,right=of T3] (T3g) {};
\node[white state,right=of T3g] (T4) {};
\node[empty state,right=of T4] (T4g) {};
\node[white state,right=of T4g] (T5) {};
\node[empty state,right=of T5] (T5g) {};
\node[white state,right=of T5g] (T6) {};

\node[white state] (S1) [below left=of T1] {};
\node[empty state,right=of S1] (S1g) {};
\node[grey state,right=of S1g] (S2) {};
\node[empty state,right=of S2] (S2g) {};
\node[grey state,right=of S2g] (S3) {};
\node[empty state,right=of S3] (S3g) {};
\node[white state,right=of S3g] (S4) {};
\node[empty state,right=of S4] (S4g) {};
\node[white state,right=of S4g] (S5) {};
\node[empty state,right=of S5] (S5g) {};
\node[white state,right=of S5g] (S6) {};
\node[empty state,right=of S6] (S6g) {};
\node[white state,right=of S6g] (S7) {};

\node[grey state] (R1) [below =of S1] {};
\node[empty state,right=of R1] (R1g) {};
\node[grey state,right=of R1g] (R2) {};
\node[empty state,right=of R2] (R2g) {};
\node[grey state,right=of R2g] (R3) {};
\node[empty state,right=of R3] (R3g) {};
\node[white state,right=of R3g] (R4) {};
\node[empty state,right=of R4] (R4g) {};
\node[white state,right=of R4g] (R5) {};
\node[empty state,right=of R5] (R5g) {};
\node[white state,right=of R5g] (R6) {};
\node[empty state,right=of R6] (R6g) {};
\node[white state,right=of R6g] (R7) {};

\node[grey state] (B1) [below =of R2] {};
\node[empty state,right=of B1] (B1g) {};
\node[white state,right=of B1g] (B2) {};
\node[empty state,right=of B2] (B2g) {};
\node[white state,right=of B2g] (B3) {};
\node[empty state,right=of B3] (B3g) {};
\node[white state,right=of B3g] (B4) {};
\node[empty state,right=of B4] (B4g) {};
\node[white state,right=of B4g] (B5) {};

\path[every node/.style={sloped,anchor=south,color=gray,auto=false,font=\tiny}]

(T1) edge[->,thick]   node[]  {}   (S1)
(T1)  edge[->,thick]   node[]  {}   (S2)
(T1)  edge[->,thick]   node[]  {}   (S3)
(T2)  edge[->,thick]   node[]  {}   (S1)
(T2)  edge[->,color=black, ultra thick]   node[]  {}   (S2)
(T2)  edge[->,color=black, ultra thick]   node[]  {}   (S3)
(T2)  edge[->,thick]   node[]  {}   (S4)
(T3)  edge[->,thick]   node[]  {}   (S2)
(T3)  edge[->,thick]   node[]  {}   (S3)
(T3)  edge[->,thick]   node[]  {}   (S4)
(T3)  edge[->,thick]   node[]  {}   (S5)
(T4)  edge[->,thick]   node[]  {}   (S3)
(T4)  edge[->,thick]   node[]  {}   (S4)
(T4)  edge[->,thick]   node[]  {}   (S5)
(T4)  edge[->,thick]   node[]  {}   (S6)
(T5)  edge[->,thick]   node[]  {}   (S4)
(T5)  edge[->,thick]   node[]  {}   (S5)
(T5)  edge[->,thick]   node[]  {}   (S6)
(T5)  edge[->,thick]   node[]  {}   (S7)
(T6)  edge[->,thick]   node[]  {}   (S5)
(T6)  edge[->,thick]   node[]  {}   (S6)
(T6)  edge[->,thick]   node[]  {}   (S7)

(S1) edge[->,thick]   node[]  {}   (R1)
(S1) edge[->,thick]   node[]  {}   (R2)
(S2) edge[->,color=black, ultra thick]   node[]  {}   (R1)
(S2) edge[->,color=black, ultra thick]   node[]  {}   (R2)
(S2) edge[->,color=black, ultra thick]   node[]  {}   (R3)
(S3) edge[->,color=black, ultra thick]   node[]  {}   (R2)
(S3) edge[->,color=black, ultra thick]   node[]  {}   (R3)
(S3) edge[->,thick]   node[]  {}   (R4)
(S4) edge[->,thick]   node[]  {}   (R3)
(S4) edge[->,thick]   node[]  {}   (R4)
(S4) edge[->,thick]   node[]  {}   (R5)
(S5) edge[->,thick]   node[]  {}   (R4)
(S5) edge[->,thick]   node[]  {}   (R5)
(S5) edge[->,thick]   node[]  {}   (R6)
(S6) edge[->,thick]   node[]  {}   (R5)
(S6) edge[->,thick]   node[]  {}   (R6)
(S6) edge[->,thick]   node[]  {}   (R7)
(S7) edge[->,thick]   node[]  {}   (R6)
(S7) edge[->,thick]   node[]  {}   (R7)

(R1) edge[->,color=black, ultra thick]   node[]  {}   (B1)
(R3) edge[->,color=black, ultra thick]   node[]  {}   (B1)
(R2) edge[->,thick]   node[]  {}   (B2)
(R4) edge[->,thick]   node[]  {}   (B2)
(R3) edge[->,thick]   node[]  {}   (B3)
(R5) edge[->,thick]   node[]  {}   (B3)
(R4) edge[->,thick]   node[]  {}   (B4)
(R6) edge[->,thick]   node[]  {}   (B4)
(R5) edge[->,thick]   node[]  {}   (B5)
(R7) edge[->,thick]   node[]  {}   (B5)
;
\end{scope}

\end{tikzpicture}
\end{minipage}
\begin{minipage}{0.45\textwidth}
\centering
 \begin{tikzpicture}
\begin{scope}[node distance=1.2cm and 0.03cm]

\tikzstyle{every initial by arrow}=[initial text=]
\tikzstyle{every state}=[circle,fill=none,draw=black,text=black,inner sep=0pt,minimum size=2.5mm]
\tikzstyle{white state}=[circle,fill=none,draw=black,text=black,inner sep=0pt,minimum size=2.5mm]
\tikzstyle{empty state}=[fill=none,inner sep=0pt,minimum size=2.5mm,opacity=0.0]
\tikzstyle{grey state}=[circle,fill=black,draw=black,text=black,inner sep=0pt,minimum size=2.5mm]

\node[white state] (T1) {};
\node[empty state,right=of T1] (T1g) {};
\node[grey state,right=of T1g] (T2) {};
\node[empty state,right=of T2] (T2g) {};
\node[white state,right=of T2g] (T3) {};
\node[empty state,right=of T3] (T3g) {};
\node[white state,right=of T3g] (T4) {};
\node[empty state,right=of T4] (T4g) {};
\node[white state,right=of T4g] (T5) {};
\node[empty state,right=of T5] (T5g) {};
\node[white state,right=of T5g] (T6) {};

\node[white state] (S1) [below left=of T1] {};
\node[empty state,right=of S1] (S1g) {};
\node[grey state,right=of S1g] (S2) {};
\node[empty state,right=of S2] (S2g) {};
\node[grey state,right=of S2g] (S3) {};
\node[empty state,right=of S3] (S3g) {};
\node[white state,right=of S3g] (S4) {};
\node[empty state,right=of S4] (S4g) {};
\node[white state,right=of S4g] (S5) {};
\node[empty state,right=of S5] (S5g) {};
\node[white state,right=of S5g] (S6) {};
\node[empty state,right=of S6] (S6g) {};
\node[white state,right=of S6g] (S7) {};

\node[grey state] (R1) [below =of S1] {};
\node[empty state,right=of R1] (R1g) {};
\node[grey state,right=of R1g] (R2) {};
\node[empty state,right=of R2] (R2g) {};
\node[grey state,right=of R2g] (R3) {};
\node[empty state,right=of R3] (R3g) {};
\node[white state,right=of R3g] (R4) {};
\node[empty state,right=of R4] (R4g) {};
\node[white state,right=of R4g] (R5) {};
\node[empty state,right=of R5] (R5g) {};
\node[white state,right=of R5g] (R6) {};
\node[empty state,right=of R6] (R6g) {};
\node[white state,right=of R6g] (R7) {};

\node[grey state] (B1) [below =of R2] {};
\node[empty state,right=of B1] (B1g) {};
\node[white state,right=of B1g] (B2) {};
\node[empty state,right=of B2] (B2g) {};
\node[white state,right=of B2g] (B3) {};
\node[empty state,right=of B3] (B3g) {};
\node[white state,right=of B3g] (B4) {};
\node[empty state,right=of B4] (B4g) {};
\node[white state,right=of B4g] (B5) {};

\path[every node/.style={sloped,anchor=south,color=gray,auto=false,font=\tiny}]

(T1) edge[->,thick]   node[]  {}   (S1)
(T1)  edge[->,thick]   node[]  {}   (S2)
(T1)  edge[->,thick]   node[]  {}   (S3)
(T2)  edge[->,thick]   node[]  {}   (S1)
(T2)  edge[->,color=black, ultra thick]   node[]  {}   (S2)
(T2)  edge[->,color=black, ultra thick]   node[]  {}   (S3)
(T2)  edge[->,thick]   node[]  {}   (S4)
(T3)  edge[->,thick]   node[]  {}   (S2)
(T3)  edge[->,thick]   node[]  {}   (S3)
(T3)  edge[->,thick]   node[]  {}   (S4)
(T3)  edge[->,thick]   node[]  {}   (S5)
(T4)  edge[->,thick]   node[]  {}   (S3)
(T4)  edge[->,thick]   node[]  {}   (S4)
(T4)  edge[->,thick]   node[]  {}   (S5)
(T4)  edge[->,thick]   node[]  {}   (S6)
(T5)  edge[->,thick]   node[]  {}   (S4)
(T5)  edge[->,thick]   node[]  {}   (S5)
(T5)  edge[->,thick]   node[]  {}   (S6)
(T5)  edge[->,thick]   node[]  {}   (S7)
(T6)  edge[->,thick]   node[]  {}   (S5)
(T6)  edge[->,thick]   node[]  {}   (S6)
(T6)  edge[->,thick]   node[]  {}   (S7)

(S1) edge[->,thick]   node[]  {}   (R1)
(S1) edge[->,thick]   node[]  {}   (R2)
(S2) edge[->,color=black, ultra thick]   node[]  {}   (R1)
(S2) edge[->,color=black, ultra thick]   node[]  {}   (R2)
(S2) edge[->,color=black, ultra thick]   node[]  {}   (R3)
(S3) edge[->,color=black, ultra thick]   node[]  {}   (R2)
(S3) edge[->,color=black, ultra thick]   node[]  {}   (R3)
(S3) edge[->,thick]   node[]  {}   (R4)
(S4) edge[->,thick]   node[]  {}   (R3)
(S4) edge[->,thick]   node[]  {}   (R4)
(S4) edge[->,thick]   node[]  {}   (R5)
(S5) edge[->,thick]   node[]  {}   (R4)
(S5) edge[->,thick]   node[]  {}   (R5)
(S5) edge[->,thick]   node[]  {}   (R6)
(S6) edge[->,thick]   node[]  {}   (R5)
(S6) edge[->,thick]   node[]  {}   (R6)
(S6) edge[->,thick]   node[]  {}   (R7)
(S7) edge[->,thick]   node[]  {}   (R6)
(S7) edge[->,thick]   node[]  {}   (R7)

(R1) edge[->,color=black, ultra thick]   node[]  {}   (B1)
(R3) edge[->,color=black, ultra thick]   node[]  {}   (B1)
(R2) edge[->,thick]   node[]  {}   (B2)
(R4) edge[->,thick]   node[]  {}   (B2)
(R3) edge[->,thick]   node[]  {}   (B3)
(R5) edge[->,thick]   node[]  {}   (B3)
(R4) edge[->,thick]   node[]  {}   (B4)
(R6) edge[->,thick]   node[]  {}   (B4)
(R5) edge[->,thick]   node[]  {}   (B5)
(R7) edge[->,thick]   node[]  {}   (B5)
;
\end{scope}

\end{tikzpicture}
\end{minipage}
\caption{Contrastive Explanation of Casual Graphs Using the Difference Condition}
\label{fig:contrast-causal-graph}
\end{figure}

As another example, consider the abstract causal graphs in Figure~\ref{fig:contrast-causal-graph}, which are larger but still trivial by AI standards. The graph on the left is of the factual case, while the graph on the right is of the foil. The dark nodes and thick vertices are what is \emph{different} between the two graphs --- the light nodes and thin vertices are the same.  It shows that changing the second input variable produces a difference value for the first output variable. When asking a contrastive \emph{why}-question about these two causal graphs, a full explanation consists of explaining the causes in both graphs. However, the difference condition merely needs us to describe those darker regions, which is a simpler task. Importantly, note that the differences are symmetric: thus, we need only explain one set of differences, and therefore, we need not explain both graphs independently. It is not a stretch to argue that contrastive questions will involved two highly similar causal graphs. For example, one is likely to question why a bird is not classified as some other type of bird, and far less likely to question why a bird is not classified as a crocodile.
\end{extended}

\subsubsection{Abnormality}
\label{sec:cognitive-processes:abnormality}

Related to the idea of contrastive explanation, \citeA{hilton1986knowledge} propose the \emph{abnormal conditions model}, based on observations from legal theorists \citeA{hart1985causation}.  \citeauthor{hilton1986knowledge}  argue that \emph{abnormal} events play a key role in causal explanation. They argue that, while statistical notions of co-variance are not the only method employed in everyday explanations, the basic idea that people select unusual events to explain is valid. Their theory states that explainers use their perceived background knowledge with explainees to select those conditions that are considered \emph{abnormal}. They give the example of asking why the Challenger shuttle exploded in 1986 (rather than not exploding, or perhaps why most other shuttles do not explode). The explanation that it exploded ``because of faulty seals'' seems like a better explanation than ``there was oxygen in the atmosphere''.  The abnormal conditions model accounts for this by noting that an explainer will reason that oxygen is present in the atmosphere when all shuttles launch, so this is not an abnormal condition. On the other hand, most shuttles to not have faulty seals, so this contributing factor was a necessary yet abnormal event in the Challenger disaster.

The abnormal conditions model has been backed up by subsequent experimental studies, such as those by \citeA{mcclure1998goals}, \citeA{mcclure2003role}, and \citeA{hilton2005course}, and more recently, \citeA{samland2014social}, who show that a variety of non-statistical measures are valid foils.

\subsubsection{Intentionality and Functionality}
\label{sec:cognitive-processes:intentionality-and-functionality}

Other features of causal chains have been demonstrated to be more important than abnormality.

\citeA{hilton2005course} investigate the claim from legal theorists \citeA{hart1985causation} that intentional action takes priority of non-intentional action in opportunity chains. Their perspective builds on the abnormal conditions model, noting that there are two important contrasts in explanation selection: (1) normal vs.\ abnormal; and (2) intentional vs.\ non-intentional. They argue further that causes will be ``traced through'' a proximal (more recent) abnormal condition if there is a more distal (less recent) event that is intentional. For example, to explain why someone died, one would explain that the poison they ingested as part of a meal was the cause of death; but if the poison as shown to have been deliberately placed in an attempt to murder the victim, the intention of someone to murder the victim receives priority. In their experiments, they gave participants different opportunity chains in which a proximal abnormal cause was an intentional human action, an unintentional human action, or a natural event, depending on the condition to which they were assigned. For example, a cause of an accident was ice on the road, which was enabled by either someone deliberative spraying the road, someone unintentionally placing water on the road, or water from a storm. Participants were asked to rate the explanations. Their results showed that: (1) participants rated intentional action as a better explanation than the other two causes, and non-intentional action better than natural cases; and (2) in opportunity chains, there is little preference for proximal over distal events if two events are of the same type (e.g.\ both are natural events) --- both are seen as necessary.


\citeA{lombrozo2010causal} argues further that this holds for \emph{functional} explanations in general; not just intentional action. For instance, citing the functional reason that an object exists is preferred to mechanistic explanations.

\subsubsection{Necessity, Sufficiency and Robustness}

Several authors \cite{lipton1990contrastive,lombrozo2010causal,woodward2006sensitive} argue that \emph{necessity} and \emph{sufficiency} are strong criteria for preferred explanatory causes. \citeA{lipton1990contrastive} argues that necessary causes are preferred to sufficient causes. For example, consider mutations in the DNA of a particular species of beetle that cause its wings to grow longer than normal when kept in certain temperatures. Now, consider that there is two such mutations, $M_1$ and $M_2$, and either is sufficient to cause the mutation. To contrast with a beetle whose wings would not change, the explanation of temperature is preferred to either of the mutations $M_1$ or $M_2$, because neither $M_1$ nor $M_2$ are individually necessary for the observed event; merely that \emph{either} $M_1$ or $M_2$. In contrast, the temperature is necessary, and is preferred, even if we know that the cause was $M_1$.

\citeA{woodward2006sensitive} argues that sufficiency is another strong criteria, in that people prefer causes that bring about the effect without any other cause. This should not be confused with sufficiency in the example above, in which either mutation $M_1$ or $M_2$ is sufficient in combination with temperature. \citeauthor{woodward2006sensitive}'s argument applies to uniquely sufficient causes, rather than cases in which there are multiple sufficient causes. For example, if it were found that are third mutation $M_3$ could cause longer wings irrelevant of the temperature, this would be preferred over temperature plus another mutation. This is related to the notation of \emph{simplicity} discussed in Section~\ref{sec:cognitive-processes:evaluation:coherence}.

Finally, several authors \cite{lombrozo2010causal,woodward2006sensitive}  argue that \emph{robustness} is also a criterion for explanation selection, in which the extend to which a cause $C$ is considered robust is whether the effect $E$ would still have occurred if conditions other than $C$ were somewhat different. Thus, a cause $C_1$ that holds only in specific situations has less explanatory value than cause $C_2$, which holds in many other situations.

\subsubsection{Responsibility}
\label{sec:cognitive-processes:responsibility}

The notions of \emph{responsibility} and \emph{blame} are relevant to causal selection, in that an event considered more responsible for an outcome is likely to be judged as a better explanation than other causes. In fact, it relates closely to necessity, as responsibility aims to place a measure of `degree of necessity' of causes. An event that is fully responsible outcome for an event is a necessary cause. 

\citeA{chockler2004responsibility} modified the structural equation model proposed by \citeA{halpern2005causes-part-I} (see Section~\ref{sec:philosophical-foundations:causality}) to define responsibility of an outcome. Informally, they define the responsibility of cause $C$ to event $E$ under a situation based on the minimal number of changes required to the situation to make event $E$ no longer occur. If $N$ is the minimal number of changes required, then the responsibility of $C$ causes $E$ is $\frac{1}{N+1}$. If $N=0$, then $C$ is fully responsible. Thus, one can see that an event that is considered more responsible than another requires less changes to prevent $E$ than the other. 

While several different cognitive models of responsibility attribution have been proposed (c.f.\ \cite{hilton2005counterfactuals,lagnado2008judgments}), I focus on the model of \citeA{chockler2004responsibility} because, as far I am aware aware, experimental evaluation of the model shows it to be stronger than existing models \cite{gerstenberg2010spreading}, and because it is a formal model that is more readily adopted in artificial intelligence.

The structural model approach defines the responsibility of \emph{events}, rather than individuals or groups, but one can see that it can be used in group models as well. \citeA{gerstenberg2010spreading} show that the model has strong predictive power at attributing responsibility to individuals in groups. They ran a set of experiments in which participants played a simple game in teams in which each individual was asked to count the number of triangles in an image, and teams won or lost depending on how accurate their \emph{collective} counts were. After the game, participants rated the responsibility of each player to the outcome. Their results showed that the modified structural equation model \citeA{chockler2004responsibility} was more accurate at predicting participants outcomes than simple counterfactual model and the so-called \emph{Matching Model}, in which the responsibility is defined as the degree of deviation to the outcome; in the triangle counting game, this would be how far off the individual was to the actual number of triangles.

\subsubsection{Preconditions, Failure, and Intentions}

An early study into explanation selection in cases of more than one cause was undertaken by \citeA{leddo1984conjunctive}. They conducted studies asking people to rate the probability of different factors as causes of events. As predicted by the intention/goal-based theory, goals were considered better explanations than relevant preconditions. However, people also rated conjunctions of preconditions and goals as better explanations of why the event occurred. For example, for the action ``\emph{Fred went to the restaurant''}, participants rated explanations such as ``\emph{Fred was hungry}'' more likely than ``\emph{Fred had money in his pocket}'', but further ``\emph{Fred was hungry and had money in his pocket}'' as an even more likely explanation, despite the fact the cause itself is less likely (conjoining the two probabilities). This is consistent with the well-known \emph{conjunction fallacy} \cite{tversky1983extensional}, which shows that people sometimes estimate the probability of the conjunction of two facts higher than either of the individual fact if those two facts are representative of prior beliefs.

However,  \citeA{leddo1984conjunctive} further showed that for \emph{failed or uncompleted} actions, just one cause (goal or precondition) was considered a better explanation, indicating that failed actions are explained differently. This is consistent with physical causality explanations \cite{lombrozo2009explanation}.  \citeauthor{leddo1984conjunctive} argue that  to explain an action, people combine their knowledge of the particular situation with a more general understanding about causal relations.  \citeA{lombrozo2010causal} argues similarly that this is because failed actions are not goal-directed, because people do not intend to fail. Thus, people prefer \emph{mechanistic} explanations for failed actions, rather than explanations that cite intentions.

\citeA{mcclure1997you} and \citeA{mcclure2001rich} found that people tend to assign a higher probability of conjoined goal and precondition for a successful action, even though they prefer the goal as the best explanation, \emph{except} in extreme/unlikely situations; that is, when the  precondition is unlikely to be true. They argue that is largely due to the (lack of) \emph{controllability} of unlikely actions. That is, extreme/unlikely events are judged to be harder to control, and thus actors would be less likely to intentionally select that action \emph{unless} the unlikely opportunity presented itself. However, for normal and expected actions, participants preferred the goal alone as an explanation instead of the goal and precondition.

 In a follow-up study, \citeA{mcclure1998goals} looked at explanations of obstructed vs.\ unobstructed events, in which an event is obstructed by its precondition being false; for example, ``\emph{Fred wanted a coffee, but did not have enough money to buy one''} as an explanation for why Fred failed to get a coffee. They showed that while goals are important to both, for obstructed events, the precondition becomes more important than for unobstructed events. 


\begin{extended}
\citeA{boonzaier2005distinguishing} go further and examine the potential overlap between beliefs and preconditions. They demonstrate that while preconditions make actions possible, they also motivate behaviour because people adopt new intentions to be able to execute those actions; making the precondition a proximal intention.
In essence, making preconditions true in order to execute a desired action (one that achieves part of a desire) becomes a proximal intention, and therefore is an \emph{indirect} cause for forming intentions. They further demonstrate that both environmental preconditions and preconditions about abilities mediate the selection of actions relative to goals. While \citeA[p.\ 206]{mcclure2002goal} had previously discussed this: ``\emph{Of course, goals may sometimes serve as a precondition to an action}'', \citeauthor{boonzaier2005distinguishing} were the first to investigate the overlap. Previously, ``\emph{the knowledge structure approach treats goals and preconditions as distinct categories}'' \citeA[p.\ 206]{mcclure2002goal}.

\citeauthor{boonzaier2005distinguishing}'s \cite{boonzaier2005distinguishing} studies show that people rate explanations using goals better than those using preconditions when people have the relevant beliefs and desires, but more importantly, that an actor's belief about their ability to reach their goals influences the choice of their proximal intentions. They further show that observers adopt a theory of mind about actors, and will attribute particular goals to actors who cannot achieve those goals, provided that the actor believes they can; even if the observers do not themselves believe that goal is achievable. Thus, desires and beliefs are relevant for predicting the proximate intentions of people, but preconditions also play a role in predicting the intentions \emph{and} actions.
\end{extended}

\begin{extended}
\subsubsection{Competing Goals}

\citeA{mcclure1989conjunctive} study the explanation of \emph{competing goals} as explanations for contrastive actions. For example, Fred did not go to the restaurant even though he wanted to, because he had a ticket to the football and wanted to go to the football more. They gave participants different variants of a vignette about Tom, who was waiting for a bus to town. In some variants, he boarded the bus, and in others, he did not. Participants were then asked to rate the probability of different explanations based on Tom's goal, such as (not) wanting to go to town and (not) wanting to wait for his friends; and different preconditions, such as his friends arriving soon or the bus being full. They showed that an explanation that uses the conjunction of the precondition and the goal of a \emph{competing} (contrastive) action was considered a better explanation than than precondition or goal alone; for example, Fred's possession of a ticket and desire to see a football match is a better explanation of why he went to the football instead of to the restaurant than either the ticket explanation or the desire explanation on its own.
\end{extended}

\subsection{Explanation Evaluation}
\label{sec:cognitive-processes:explanation-evaluation}

In this section, we look at work that has investigated the criteria that people use to evaluate explanations. The most important of these are: probability, simplicity, generalise, and coherence with prior beliefs.

\subsubsection{Coherence, Simplicity, and Generality}
\label{sec:cognitive-processes:evaluation:coherence}

\citeA{thagard1989explanatory} argues that coherence is a primary criteria for explanation. He proposes the \emph{Theory for Explanatory Coherence}, which specifies seven principles of how explanations relate to prior belief. He argues that these principles are foundational principles that explanations must observe to be acceptable. They capture properties such as if some set of properties \P explain some other property \Q, then all properties in \P must be coherent with \Q; that is, people will be more likely to accept explanations if they are consistent with their prior beliefs. Further, he contends that all things being equal, simpler explanations --- those that cite fewer causes --- and more general explanations --- those that explain more events ---, are better explanations. The model has been demonstrated to align with how humans make judgements on explanations \cite{ranney1988explanatory}.

\citeA{read1993explanatory} tested the hypotheses from \citeauthor{thagard1989explanatory}'s theory of explanatory coherence \cite{thagard1989explanatory} that people prefer simpler and more general explanations. Participants were asked to rate the probability and the `quality' of explanations with different numbers of causes. They were given stories containing several events to be explained, and several different explanations. For example, one story was about Cheryl, who is suffering from three medical problems: (1) weight gain; (2) fatigue; and (3) nausea. Different participant groups were given one of three types of explanations: (1) \emph{narrow}: one of Cheryl having stopped exercising (weight gain), has mononucleosis (explains fatigue), or a stomach virus (explains nausea); (2) \emph{broad}: Cheryl is pregnant (explains all three); or (3) \emph{conjunctive}: all three from item 1 as the same time. As predicted, participants preferred simple explanations (pregnancy) with less causes than more complex ones (all three conjunctions), and participants preferred explanations that explained more events.

\subsubsection{Truth and Probability}
\label{sec:cognitive-processes:probability}

\begin{changed}
Probability has two facets in explanation: the probability of the explanation being true; and the \emph{use} of probability in an explanation. Neither has a much importance as one may expect.

The use of statistical relationships to explain events is considered to be unsatisfying on its own. This is because people desire \emph{causes} to explain events, not associative relationships. \citeA{josephson1996abductive} give the example of a bag full of red balls. When selecting a ball randomly from the bag, it must be red, and one can ask: ``Why is this ball red?''. The answer that uses the statistical generalisation ``Because all balls in the bag are red'' is not a good explanation, because it does not explain why that particular ball is red. A better explanation is someone painted it red. However, for the question: ``Why did we observe a red ball coming out of the bag'', it is a good explanation, because having only red balls in the bag does cause us to select a red one. \citeauthor{josephson1996abductive} highlight that the difference between explaining the \emph{fact observed} (the ball is red) and explaining \emph{the event of observing the fact} (a red ball was selected).  To explain instances via statistical generalisations, we need to explain the \emph{causes} of those generalisations too, not the generalisations themselves. If the reader is not convinced, consider my own example: a student coming to their teacher to ask why they only received 50\% on an exam. An explanation that most students scored around 50\% is not going to satisfy the student. Adding a cause for why most students only scored 50\% would be an improvement. Explaining to the student why they specifically received 50\% is even better, as it explains the cause of the instance itself.
\end{changed}

The truth of likelihood of an explanation is considered an important criteria of a good explanation. However, \citeA{hilton1996mental} shows that the most likely or `true'' cause is not necessarily the \emph{best} explanation. Truth conditions\footnote{We use the term \emph{truth condition} to refer to facts that are either true or considered likely by the explainee.} are a necessary but not sufficient criteria for the generation of explanations. While a true or likely cause is one attribute of a good explanation, tacitly implying that the most probable cause is always the best explanation is incorrect. As an example, consider again the explosion of the Challenger shuttle (Section~\ref{sec:cognitive-processes:abnormality}), in which a faulty seal was argued to be a better explanation than oxygen in the atmosphere. This is despite the fact the the `seal' explanation is a likely but not known cause, while the `oxygen' explanation is a known cause. \citeauthor{hilton1996mental} argues that this is because the fact that there is oxygen in the atmosphere is \emph{presupposed}; that is, the explainer assumes that the explainee already knows this.

\citeA{mcclure2002goal} also challenges the idea of probability as a criteria for explanations. Their studies found that people tend not to judge the quality of explanations around their probability, but instead around their so-called \emph{pragmatic influences} of causal behaviour. That is, people judge explanations on their usefulness, relevance, etc., including via Grice's maxims of conversation \cite{grice1975logic} (see Section~\ref{sec:social-explanation:grice} for a more detailed discussion of this). This is supported  by experiments such as \citeA{read1993explanatory} cited above, and the work from \citeA{tversky1983extensional} on the conjunction fallacy.

\citeA{lombrozo2007simplicity} notes that the experiments on generality and simplicity performed by \citeA{read1993explanatory} cannot rule out that participants selected simple explanations because they did not have probability or frequency information for events. \citeauthor{lombrozo2007simplicity} argues that if participants assumed that the events of stopping exercising, having mononucleosis, having a stomach virus, and being pregnant are all equally likely, then the probability of the conjunction of any three is much more unlikely than any one combined.
To counter this, she investigated the influence that probability has on explanation evaluation, in particular, when simpler explanations are less probable than more complex ones. Based on a similar experimental setup to that of \citeA{read1993explanatory}, \citeauthor{lombrozo2007simplicity} presented experimental participants with information about a patient with several symptoms that could be explained by one cause or several separate causes. In some setups, base rate information about each disease was provided, in which the conjunction of the separate causes was more likely than the single (simpler) cause. Without base-rate information, participants selected the most simple (less likely) explanations. When base-rate information was included, this still occurred, but the difference was less pronounced. However, the likelihood of the conjunctive scenario had to be \emph{significantly more likely for it to be chosen}.  \citeauthor{lombrozo2007simplicity}'s final experiment showed that this effect was reduced again if participants were \emph{explicitly} provided with the joint probability of the two events, rather than in earlier experiments in which they were provided separately.

\citeA{preston2005explanations} show that the value that people assign to their own beliefs -- both in terms of probability and personal relevance -- correspond with the \emph{explanatory power} of those beliefs.  Participants were each given a particular `belief' that is generally accepted by psychologists, but mostly unknown in the general public, and were then allocated to three conditions: (1) the \emph{applications} condition, who were asked to list observations that the belief could explain; (2) the \emph{explanations} condition, who were asked to list observations that could explain the belief (the inverse to the previous condition); and (3) a control condition who did neither. Participants were then asked to consider the probability of that belief being true, and to assign their perceived \emph{value} of the belief to themselves and society in general. Their results show that people in the applications and explanations condition both assigned a higher probability to the belief being true, demonstrating that if people link beliefs to certain situations, the perceived probability increased. However, for value, the results were different: those in the applications condition assigned a higher value than the other two conditions, and those in the explanations condition assigned a lower value than the other two conditions. This indicates that people assign higher values to beliefs that explain observations, but a lower value to beliefs that can be explained by \emph{other} observations.

\begin{changed}
 \citeA{kulesza2013too} investigate the balance between soundness and completeness of explanation. They investigated explanatory debugging of machine learning algorithms making personalised song recommendations. By using progressively simpler models with less features, they trained a recommender system to give less correct recommendations.  Participants were given recommendations for songs on a music social media site, based on their listening history, and were placed into one of several treatments. Participants in each treatment would be given a different combination of soundness and completeness, where soundness means that the explanation is correct and completeness means that all of the underlying causes are identified.  For example, one treatment had low soundness but high completeness, while another had medium soundness and medium completeness. Participants were given a list of recommended songs to listen to, along with the (possibly unsound and incomplete) explanations, and were subsequently asked why the song had been recommended. The participants' mental models were measured. The results show that sound and complete models were the best for building a correct mental model, but at the expense of cost/benefit. Complete but unsound explanations improved the participants' mental models more than soundness, and gave a better perception of cost/benefit, but reduced trust. Sound but incomplete explanations were the least preferred, resulting in higher costs and more requests for clarification. Overall, \citeauthor{kulesza2013too} concluded that completeness was more important than soundness. From these results, \citeA{kulesza2015principles} list three principles for explainability: (1) \emph{Be sound}; (2) \emph{Be complete}; but (3) \emph{Don't overwhelm}. Clearly, principles 1 and 2 are at odds with principle 3, indicating that careful design must be put into explanatory debugging systems.
\end{changed}

\subsubsection{Goals and Explanatory Mode}

\citeA{vasilyeva2015goals} show that the goal of explainer is key in how the evaluated explanations, in particular, in relation to the \emph{mode} of explanation used (i.e.\ material, formal, efficient, final). In their experiments, they gave participants different tasks with varying goals. For instance, some participants were asked to assess the causes behind some organisms having certain traits (efficient), others were asked to categorise organisms into groups (formal), and the third group were asked for what reason organisms would have those traits (functional). They provided explanations using different modes for parts of the tasks and then asked participants to rate the `goodness' of an explanation provided to them. Their results showed that the goals not only shifted the focus of the questions asked by participants, but also that participants preferred modes of explanation that were more congruent with the goal of their task. This is further evidence that being clear about the question being asked is important in explanation.

\subsection{Cognitive Processes and XAI}
\label{sec:cognitive-processes:AI}

This section presents some ideas on how the work on the cognitive processes of explanation affects researchers and practitioners in XAI. 

\begin{changed}
The idea of explanation selection is not new in XAI. Particularly in machine learning, in which models have many features, the problem is salient. Existing work has primarily looked at selecting which features in the model were important for a decision, mostly built on local explanations \cite{robnik2008explaining,baehrens2010explain,ribeiro2016should} or on information gain \cite{kulesza2013too,kulesza2015principles}.
However, as far as the authors are aware, there are currently no studies that look at the cognitive biases of humans as a way to select explanations from a set of causes.
\end{changed}

\subsubsection{Abductive Reasoning}

\begin{changed}
Using abductive reasoning to generate explanations has a long history in artificial intelligence \cite{levesque1989knowledge}, aiming to solve problems such as fault diagnosis \cite{poole1989normality}, plan/intention recognition \cite{charniak1991probabilistic}, and generalisation in learning \cite{mitchell1986explanation}. Findings from such work has parallels with many of the results from cognitive science/psychology outlined in this section. \citeA{leake1995abduction} provides an excellent overview of the challenges of abduction for everyday explanation, and summarises work that addresses these. He notes three of the main tasks that an abductive reasoner must perform are: (1) what to explain about a given situation (determining the question); (2) how to generate explanations (abductive reasoning); and (3) how to evaluate the ``best'' explanation (explanation selection and evaluation). He stresses that determining the goal of the explanation is key to providing a good explanation; echoing the social scientists' view that the explainee's question is important, and that such questions are typically focused on anomalies or surprising observations.
\end{changed}

The work from \citeauthor{rehder2003causal} \cite{rehder2003causal,rehder2006similarity} and \citeA{lombrozo2012explanation} show that that explanation is good for learning and generalisation. This is interesting and relevant for XAI, because it shows that individual users should require less explanation the more they interact with a system. First, because they will construct a better mental model of the system and be able to generalise its behaviour (effectively learning its model). Second, as they see more cases, they should become less surprised by abnormal phenomena, which as noted in Section~\ref{sec:cognitive-processes:abnormality}, is a primary trigger for requesting explanations. An intelligent agent that presents --- unprompted -- an explanation alongside every decision, runs a risk of providing explanations that become less needed and more distracting over time.

\begin{changed}
The work on inherent vs.\ extrinsic features (Section~\ref{sec:cognitive-processes:abductive-reasoninig:inherence}) is relevant for many AI applications, in particular classification tasks. In preliminary work, \citeA{bekele2018human} use the inherence bias \cite{cimpian2014inherence} to explain person identification in images. Their re-identification system is tasked with determining whether two images contain the same person, and uses inherent features such as age, gender, and hair colour, as well as extrinsic features such as clothing or wearing a backpack. Their explanations use the inherence bias with the aim of improving the acceptability of the explanation. In particular, when the image is deemed to be of the same person, extrinsic properties are used, while for different people, intrinsic properties are used. This work is preliminary and has not yet been evaluated, but it is an excellent example of using cognitive biases to improve explanations.
\end{changed}

\subsubsection{Mutability and Computation}

Section~\ref{sec:cognitive-processes:mutability} studies the heuristics that people use to discount some events over others during mental simulation of causes. This is relevant to some areas of explainable AI because, in the same way that people apply these heuristics to more efficiently search through a causal chain, so to can these heuristics be used to more efficiently find causes, while still identifying causes that a human explainee would expect.

The notions of causal temporality and responsibility would be reasonably straightforward to capture in many models, however, if one can capture concepts such as abnormality, responsibility intentional, or controllability in models, this provides further opportunities. 

\subsubsection{Abnormality}
\label{sec:cognitive-processes:AI:abnormality}

Abnormality clearly plays a role in explanation and interpretability. For explanation, it serves as a trigger for explanation, and is a useful criteria for explanation selection. For interpretability, it is clear that `normal' behaviour will, on aggregate, be judged more explainable than abnormal behaviour.

Abnormality is a key criteria for explanation selection, and as such, the ability to identify abnormal events in causal chains could improve the explanations that can be supplied by an explanatory agent. While for some models, such as those used for probabilistic reasoning, identifying abnormal events would be straightforward, and for others, such as normative systems, they are `built in', for other types of  models, identifying abnormal events could prove difficult but valuable.

One important note to make is regarding abnormality and  its application to ``\emph{non-contrastive}'' why--questions. As noted in Section~\ref{sec:philosophical-foundations:AI:contrastive-explanation}, questions of the form  ``\emph{Why \P?}'' may have an implicit foil, and determining this can improve explanation. In some cases, normality could be used to mitigate this problem. That is, in the case of ``\emph{Why \P?}'', we can interpret this as ``\emph{Why \P rather than the normal case \Q?}'' \cite{hilton1990conversational}. For example, consider the application of assessing the risk of glaucoma \cite{chan2002comparison}. Instead of asking why they were given a positive diagnosis rather than a negative diagnosis, the explanatory again could provide one or more \emph{default} foils, which would be `stereotypical' examples of people who were not diagnosed and whose symptoms were more regular with respect to the general population. Then, the question becomes why was the person diagnosed with glaucoma compared to these default stereotypical cases without glaucoma.

\subsubsection{Intentionality and Functionality}

The work discussed in Section~\ref{sec:cognitive-processes:intentionality-and-functionality} demonstrates the importance of intentionality and functionality in selecting explanations. As discussed in Section~\ref{sec:social-attribution:AI:folk-psychology}, these concepts are highly relevant to deliberative AI systems, in which concepts such as goals and intentions are first-class citizens. However, the importance of this to explanation selection rather than social attribution must be drawn out. In social attribution, folk psychological concepts such as intentions are attributed to agents to identify causes and explanations, while in this section, intentions are used as part of the cognitive process of \emph{selecting} explanations from a causal chain. Thus, even for a non-deliberative system, labelling causes as intentional could be useful. For instance, consider a predictive model in which some features represent that an intentional event has occurred. Prioritising these may lead to more intuitive explanations.

\subsubsection{Perspectives and Controllability}

The finding from \citeA{kahneman1982simulation} that perspectives change the events people mutate, discussed in Section~\ref{sec:cognitive-processes:mutability}, is important in multi-agent contexts. This implies that when explaining a particular agent's decisions or behaviour, the explanatory agent could focus on undoing actions of that particular agent, rather than others. This is  also consistent with the research on controllability discussed in Section~\ref{sec:cognitive-processes:mutability}, in that, from the perspective of the agent in question, they can only control their own actions.

In interpretability, the impact of this work is also clear: in generating explainable behaviour, with all others things being equal, agents could select actions that lead to future actions being more constrained, as the subsequent actions are less likely to have counterfactuals undone by the observer.

\subsubsection{Evaluation of Explanations}

The importance of the research outlined in Section~\ref{sec:cognitive-processes:explanation-evaluation} is clear: likelihood is not everything. While likely causes are part of good explanations, they do not strongly correlate with explanations that people find useful. The work outlined in this section provides three criteria that are at least as equally important: simplicity, generality, and coherence.

For explanation, if the goal of an explanatory agent is to provide the most likely causes of an event, then these three criteria can be used to prioritise among the most likely events. However, if the goal of an explanatory agent is to generate trust between itself and its human observers, these criteria should be considered as first-class criteria in explanation generation beside or even above likelihood. For example, providing simpler explanations that increase the likelihood that the observer both \emph{understands} and \emph{accepts} the explanation may increase trust better than giving more likely explanations.

For interpretability, similarly, these three criteria can form part of decision-making algorithms; for example, a deliberative agent may opt to select an action that is less likely to achieve its goal, if the action helps towards other goals that the observer knows about, and has a smaller number of causes to refer to.

\begin{changed}
The selection and evaluation of explanations in artificial intelligence has been studied in some detail, going back to early work on abductive reasoning, in which explanations with structural simplicity, coherence, or minimality are preferred (e.g.\ \cite{reiter1987theory,levesque1989knowledge}) and the concept of explanatory power of a set of hypotheses is defined as the set of manifestations those hypotheses account for \cite{allemang1987computational}. Other approaches use probability as the defining factor to determine the most likely explanation (e.g.\ \cite{halpern2005causes-part-II}). In addition to the cognitive biases of people to discount probability, the probabilistic approaches have the problem that such fine-grained probabilities are not always available \cite{leake1995abduction}. These selection mechanisms are context-independent and do not account for the explanations as being relevant to the question nor the explainee. 

\citeA{leake1991goal}, on the other hand, argues for goal-directed explanations in abductive reasoning that explicitly aim to reduce knowledge gaps; specifically to explain why an observed event is ``reasonable'' and to help identify faulty reasoning processes that led to it being surprising. He proposes nine \emph{evaluation dimensions} for explanations: timeliness, knowability, distinctiveness, predictive power,  causal force, independence, repairability, blockability, and desirability. Some of these correspond to evaluation criteria outlined in Section~\ref{sec:cognitive-processes:explanation-evaluation}; for example, distinctiveness notes that a cause that is surprising is of good explanatory value, which equates to the criteria of abnormality.
\end{changed}


\section{Social Explanation --- How Do People Communicate Explanations?}
\label{sec:social-explanation}

\begin{quote}
 Causal explanation is first and foremost a form of social interaction. One speaks of giving causal explanations, but not attributions, perceptions, comprehensions, categorizations, or memories. The verb to explain is a three-place predicate: \textbf{Someone} explains \textbf{something} to \textbf{someone}. Causal explanation takes the form of conversation and is thus subject to the rules of conversation. [Emphasis original] --- \citeA{hilton1990conversational}
\end{quote}

This final section looks at the communication problem in explanation --- something that has been studied little in explainable AI so far. The work outlined in this section asserts that the explanation process does not stop at just selecting an explanation, but considers that an explanation is an interaction between two roles: explainer and explainee (perhaps the same person/agent playing both roles), and that there are certain `rules' that govern this interaction.

\subsection{Explanation as Conversation}

\citeA{hilton1990conversational} presents the most seminal article on the social aspects of conversation, proposing a \emph{conversational model of explanation} based on foundational work undertaken by both himself and others. The primary argument of \citeauthor{hilton1990conversational}  is that explanation is a \emph{conversation}, and this is how it differs from causal attribution. He argues that there are two stages: the \emph{diagnosis} of causality in which the explainer determines why an action/event occurred; and the \emph{explanation}, which is the social process of conveying this to someone. The problem is  then to ``\emph{resolve a puzzle in the explainee's mind about why the event happened by closing a gap in his or her knowledge}'' \cite[p.\ 66]{hilton1990conversational}.

The conversational model argues that good social explanations must be \emph{relevant}. This means that they must answer the question that is asked ---  merely identifying causes does not provide good explanations, because many of the causes will not be relevant to the questions; or worst still, if the ``most probable'' causes are selected to present to the explainee, they will not be relevant to the question asked. The information that is communicated between explainer and explainee should conform to the general rules of \emph{cooperative} conversation \cite{grice1975logic}, including being relevant to the explainee themselves, and what they already know.

\citeA{hilton1990conversational} terms the second stage \emph{explanation presentation}, and argues that when an explainer presents an explanation to an explainee, they are engaged in a conversation. As such, they tend to follow basic rules of conversation, which \citeauthor{hilton1990conversational} argues are captured by \emph{Grice's maxims of conversation} \cite{grice1975logic}: (a) quality; (b) quantity; (c) relation; and (d) manner. Coarsely, these respectively mean: only say what you believe; only say as much as is necessary; only say what is relevant; and say it in a nice way.

  These maxims imply that the shared knowledge between explainer and explainee are \emph{presuppositions} of the explanations, and the other factors are the causes that should be explained; in short, the explainer should not explain any causes they think the explainee already knows (epistemic explanation selection).


Previous sections have presented the relevant literature about causal connection (Sections~\ref{sec:social-attribution} and \ref{sec:cognitive-processes}) and explanation selection (Sections~\ref{sec:cognitive-processes}). In the remainder of this subsection, we describe Grice's model and present related research that analyses how people select explanations relative to subjective (or social) viewpoints, and present work that supports \citeauthor{hilton1990conversational}'s conversational model of explanation \cite{hilton1990conversational}.

\subsubsection{Logic and Conversation}
\label{sec:social-explanation:grice}

\emph{Grice's maxims} \cite{grice1975logic} (or the \emph{Gricean maxims}) are a  model for how people engage in cooperative conversation. \citeauthor{grice1975logic} observes that conversational statements do not occur in isolation: they are often linked together, forming a cooperative effort to achieve some goal of information exchange or some social goal, such as social bonding. He notes then that a general principle that one should adhere to in conversation is the \emph{cooperative principle}: ``\emph{Make your conversational contribution as much as is required, at the stage at which it occurs, by the accepted purpose or direction of the talk exchange in which you are engaged}'' \cite[p.\ 45]{grice1975logic}.

For this, \citeA{grice1975logic} distinguishes four categories of maxims that would help to achieve the cooperative principle:

\begin{enumerate}

 \item \emph{Quality}: Make sure that the information is of high quality -- try to make your contribution one that is true. This contains two maxims:
 	\begin{enumerate*}
    	\item do not say things that you believe to be false; and
        \item do not say things for which you do not have sufficient evidence. 
    \end{enumerate*}

 \item \emph{Quantity}: Provide the right quantity of information. This contains two maxims:
   \begin{enumerate*}
     \item make your contribution as informative as is required; and
     \item do not make it more informative than is required.
   \end{enumerate*}
   
  \item \emph{Relation}: Only provide information that is related to the conversation. This consists of a single maxim:
   \begin{enumerate*}
   	\item Be relevant. This maxim can be interpreted as a strategy for achieving the maxim of quantity.
   \end{enumerate*}
   
  \item \emph{Manner}: Relating to \emph{how} one provides information, rather than what is provided. This consists of the `supermaxim' of `Be perspicuous', but according to Grice, is broken into `various' maxims such as:
  \begin{enumerate*}
    \item avoid obscurity of expression;
    \item avoid ambiguity;
    \item be brief (avoid unnecessary prolixity); and
    \item be orderly.
  \end{enumerate*}
\end{enumerate}

\citeA{grice1975logic} argues that for cooperative conversation, one should obey these maxims, and that people learn such maxims as part of their life experience. He further links these maxims to \emph{implicature}, and shows that it is possible to violate some maxims while still being cooperative, in order to either not violate one of the other maxims, or to achieve some particular goal, such as to \emph{implicate} something else without saying it. Irony and metaphors are examples of violating the quality maxims, but other examples, such as: Person A: ``\emph{What did you think of the food they served?}''; Person B: ``\emph{Well, it was certainly healthy}'', violates the maxim of manner, but is implying perhaps that Person B did not enjoy the food, without them actually saying so. 

Following from the claim that explanations are conversations, \citeA{hilton1990conversational} argues that explanations should follow these maxims. The quality and quantity categories present logical characterisations of the explanations themselves, while the relation and manner categories define how they explanations should be given.

\subsubsection{Relation \& Relevance in Explanation Selection}
\label{sec:social-explanation:relevance}


Of particular interest here is research to support these Gricean maxims; in particular, the related maxims of \emph{quantity} and \emph{relevance}, which together state that the speaker should only say what is necessary and relevant. In social explanation, research has shown that people select explanations to adhere to these maxims by considering the particular question being asked by the explainee, but also by giving explanations that the explainee does not already accept as being true.: To quote \citeauthor{hesslow1988problem}:

\begin{quote}
What are being selected are essentially questions, and the causal selection that follows from this is determined by the straightforward criterion of explanatory relevance. --- \cite[p.\ 30]{hesslow1988problem}
\end{quote}

In Section~\ref{sec:cognitive-processes:explanation-selection:facts-and-foils}, we saw evidence to suggest that the difference between the fact and foil for contrastive why--questions are the relevant causes for explanation. In this section, we review work on the social aspects of explanation selection and evaluation.

\paragraph{Epistemic Relevance}
\citeA{slugoski1993attribution} present evidence of Gricean maxims in explanation, and of support for the idea of explanation as conversation. They argue that the form of explanation must take into account its function as an answer to a specified why--question, and that this should take part within a conversational framework, including the context of the explainee. They gave experimental participants information in the form of a police report about an individual named George who had been charged with assault after a school fight. This information contained information about George himself, and about the circumstances of the fight. Participants were then paired with another `participant' (played by a researcher), were told that the other participant had either: (a) information about George; (2) the circumstances of the fight; or (c) neither; and were asked to answer why George had assaulted the other person. The results showed participants provided explanations that are tailored to their expectations of what the hearer already knows, selecting single causes based on abnormal factors of which  they believe the explainee is unaware; and that participants \emph{change} their explanations of the same event when presenting to explainees with differing background knowledge.

\citeA{jaspars1988mental} and \citeA{hilton1996mental} both argue that such results  demonstrate that, as well as being true or likely, a good  explanation must be relevant to both the question and to the \emph{mental model} of the explainee.  \citeA{byrne1991construction} offers a similar argument in her computational model of explanation selection, noting that humans are model-based, not proof-based, so explanations must be relevant to a model. 

\citeA{halpern2005causes-part-II} present an elegant formal model of explanation selection based on epistemic relevance. This model extends their work on structural causal models \cite{halpern2005causes-part-II}, discussed in Section~\ref{sec:philosophical-foundations:causality}. They define an explanation as a fact that, if found to be true, would constitute an actual cause of a specific event. 

Recall from Section~\ref{sec:philosophical-foundations:causality} structural causal models \cite{halpern2005causes-part-I} contain variables and functions between these variables. A \emph{situation} is a unique assignment from variables to values. \citeA{halpern2005causes-part-II} then define an \emph{epistemic state} as a set of situations, one for each possible situation that the explainee considers possible. Explaining the causes of an event then becomes providing the values for those variables that remove some situations from the epistemic state such that the cause of the event can be uniquely identified. They then further show how to provide explanations that describe the structural model itself, rather than just the values of variables, and how to reason when provided with probability distributions over events. Given a probabilistic model,  \citeauthor{halpern2005causes-part-II} formally define the \emph{explanatory power} of partial explanations. Informally, this states that explanation $C_1$ has more explanatory power explanation $C_2$ for explanandum $E$ if and only if providing $C_1$  to the explainee increases the prior probability of $E$ being true more than providing $C_2$ does.

\begin{extended}
 \citeA{halpern2005causes-part-II} note that explanations with higher explanatory power are generally more specific (contain more conjunctions) than explanations with lower explanatory power, and therefore are actually less likely to occur. They describe this as a `tension' with no obvious solution, however, as the reader will know from Section~\ref{sec:cognitive-processes:probability}, this property does not conflict with research in cognitive psychology, such as the conjunction fallacy \citeA{tversky1983extensional} and the research showing that people do not use probability as a strong criterion for the usefulness of explanations \cite{hilton1996mental} (Section~\ref{sec:cognitive-processes:probability}).
\end{extended}

\citeA{dodd1980leading} demonstrates that the perceived intention of a speaker is important in implicature. Just as leading questions in eyewitness reports can have an effect on the judgement of the eyewitness, so to it can affect explanation. They showed that the meaning and presuppositions that people infer from conversational implicatures depends heavily on the perceived intent or bias of the speaker. In their experiments, they asked participants to assess, among other things, the causes of a vehicle accident, with the account of the accident being given by different parties: a neutral bystander vs.\ the driver of the vehicle. Their results show that the bystander's information is more trusted, but also that \emph{incorrect} presuppositions are recalled as `facts' by the participants if the account was provided by the neutral source, but not the biased source; \emph{even if they observed the correct facts to begin with}. \citeauthor{dodd1980leading} argue that this is because the participants filtered the information relative to their perceived intention of the person providing the account. 

\paragraph{The Dilution Effect}
\citeA{tetlock1989accountability} investigated the effect of implicature with respect to the information presented, particularly its relevance, showing that when presented with additional, irrelevant information, people's implicatures are \emph{diluted}. They performed a series of controlled experiments in which participants were presented with information about an individual David, and were asked to make predictions about David's future; for example, what his grade point average (GPA) would be. There were two control groups and two test groups. In the control groups, people were told David spent either 3 or 31 hours studying each week (which we will call groups C3 and C31), while in the \emph{diluted} group test groups, subjects were also provided with additional irrelevant information about David (groups T3 and T31). The results showed that those in the diluted T3 group predicted a \emph{higher} GPA than those in the undiluted C3 group, while those in the diluted T31 group predicted a \emph{lower} GPA than those in the undiluted C31 group. \citeauthor{tetlock1989accountability} argued that this is because participants assumed the irrelevant information may have indeed been relevant, but its lack of support for prediction led to less extreme predictions. This study and studies on which it built demonstrate the importance of relevance in explanation.

In a further study, \citeA{tetlock1996dilution} explicitly controlled for conversational maxims, by informing one set of participants that the information displayed to them was chosen at random from the history of the individual. Their results showed that the dilution effect disappeared when conversational maxims were deactivated, providing further evidence for the dilution effect.

Together, these bodies of work and those on which they build demonstrate that Grice's maxims are indeed important in explanation for several reasons; notably that they are a good model for how people expect conversation to happen. Further, while it is clear that providing more information than necessary not only would increase the cognitive load of the explainee, but that it dilutes the effects of the information that is truly important.

\subsubsection{Argumentation and Explanation}
\label{sec:social-explanation:conversation:argument}

\citeA{antaki1992explaining} extend \citeauthor{hilton1990conversational}'s conversational model \cite{hilton1990conversational} from dialogues to \emph{arguments}. Their research shows that a majority of statements made in explanations are actually argumentative claim-backings; that is, justifying that a particular cause indeed did hold (or was thought to have held) when a statement is made. Thus, explanations are used to both report causes, but also to \emph{back claims}, which is an argument rather than just a question-answer model. They extend the conversational model to a wider class of contrast cases. As well as explaining causes, one must be prepared to defend a particular claim made in a causal explanation. Thus, explanations extend not just to the state of affairs external to the dialogue, but also to the internal attributes of the dialogue itself.


An example on the distinction between explanation and argument provided by \citeA[p. 186]{antaki1992explaining} is ``\emph{The water is hot because the central heating is on}''. The distinction lies on whether the \emph{speaker} believes that the \emph{hearer} believes that the water is hot or not. If it is believed that the speaker believes that the water is hot, then the central heating being on offers an explanation: it contrasts with a case in which the water is not hot. If the speaker believes that the hearer does not believe the water is hot, then this is an argument that the water should indeed be hot; particularly if the speaker believes that the hearer believes that the central heating is on. The speaker is thus trying to persuade the hearer that the water is hot. However, the distinction is not always so clear because explanations can have argumentative functions.

\subsubsection{Linguistic structure}

\citeA{malle2000conceptual} argue that the linguistic structure of explanations plays an important role in interpersonal explanation. They hypothesise that some linguistic devices are used not to change the reason, but to indicate \emph{perspective} and to \emph{manage impressions}. They asked experimental participants to select three negative and three positive intentional actions that they did recently that were outside of their normal routine. They then asked participants to explain why they did this, and coded the answers. Their results showed several interesting findings.

First, explanations for reasons can be provided in two different ways: \emph{marked} or \emph{unmarked}. An unmarked reason is a direct reason, while a marked reason has a \emph{mental state marker} attached. For example, to answer the question ``\emph{Why did she go back into the house}'', the explanations ``\emph{The key is still in the house}'' and ``\emph{She thinks the key is still in the house}'' both give the same reason,  but with different constructs that are used to give different impressions: the second explanation gives an impression that the explainee may not be in agreement with the actor. 

Second, people use \emph{belief} markers and \emph{desire} markers; for example, ``\emph{She thinks the key is in the house}'' and ``\emph{She wants the key to be in her pocket}'' respectively.
In general, dropping \emph{first-person markings}, that is, a speaker dropping ``\emph{I/we believe}'', is common in conversation and the listeners automatically infer that this is a belief of the speaker.  For example, ``\emph{The key is in the house}'' indicates a belief on the behalf of the speaker and inferred to mean ``\emph{I believe the key is in the house}'' \cite{malle2000conceptual}\footnote{\citeA[Chapter 4]{malle2004mind} also briefly discusses \emph{valuings} as markers, such as ``\emph{She likes}'', but notes that these are rarely dropped in reasons.}.

However, for \emph{third-person} perspective, this is not the case. The unmarked version of explanations, especially belief markers, generally imply some sort of agreement from the explainer: ``\emph{She went back in because the key is in the house}'' invites the explainee to infer that the actor and the explainer \emph{share} the belief that the key is in the house. Whereas, ``\emph{She went back in because she believes the key is in the house}'' is ambiguous --- it does not (necessarily) indicate the belief of the speaker. The reason: ``\emph{She went back in because she mistakenly believes the key is in the house}'' offers no ambiguity of the speaker's belief. 

\citeA[p.\ 169, Table~6.3]{malle2004mind} argues that different markers sit on a scale between being \emph{distancing} to being \emph{embracing}. For example, ``\emph{she mistakenly believes}'' is more distancing than ``\emph{she jumped to the conclusion'}', while ``\emph{she realises}'' is embracing. Such constructs aim not to provide different reasons, but merely allow the speaker to form impressions about themselves and the actor.

\subsection{Explanatory Dialogue}

If we accept the model of explanation as conversation, then we may ask whether there are particular dialogue structures for explanation. There has been a collection of such articles ranging from dialogues for pragmatic explanation \cite{van1977pragmatics} to definitions based on transfer of understanding \cite{vonwright1971explanation}. However, the most relevant for the problem of explanation in AI is a body of work lead largely by \citeauthor{walton2004new}.

\citeA{walton2004new} proposed a dialectical theory of explanation, putting forward similar ideas to that of \citeA{antaki1992explaining} in that some parts of an explanatory dialogue require the explainer to provide backing arguments to claims. In particular, he argues that such an approach is more suited to `everyday' or interpersonal explanation than models based on scientific explanation. He further argues that such models should be combined with ideas of \emph{explanation as understanding}, meaning that social explanation is about transferring knowledge from explainer to explainee. He proposes  a series of conditions on the dialogue and its interactions as to when and how an explainer should transfer knowledge to an explainee. 

In a follow-on paper, \citeA{walton2007dialogical} proposes a formal dialogue model called \emph{CE}, based on an earlier persuasion dialogue \cite{walton1984logical}, which defines the conditions on how a explanatory dialogue commences, rules for governing the locutions in the dialogue, rules for governing the structure or sequence of the dialogue, success rules and termination rules. 

Extending on this work further \cite{walton2007dialogical},  \citeA{walton2011dialogue} describes an improved formal dialogue system for explanation, including a set of speech act rules for practical explanation, consisting of an opening stage, exploration stage, and closing stage. In particular, this paper focuses on the closing stage to answer the question: how do we know that an explanation has `finished'?  \citeA{scriven1972concept} argues that to test someone's understanding of a topic, merely asking them to recall facts that have been told to them is insufficient --- we should also be able to answer new questions that demonstrate generalisation of and inference from what has been learnt: an \emph{examination}.

To overcome this, \citeauthor{walton2007dialogical} proposes the use of \emph{examination dialogues} \cite{walton2006examination} as a method for the explainer to determine whether the explainee has correctly understood the explanation --- that is, the explainer has a real understanding, not merely a perceived (or claimed) understanding. \citeauthor{walton2007dialogical} proposes several rules for the closing stage of the examination dialogue, including a rule for terminating due to `practical reasons', which aim to solve the problem of the \emph{failure cycle}, in which repeated explanations are requested, and thus the dialogue does not terminate.

\citeA{arioua2015formalizing} formalise \citeauthor{walton2011dialogue}'s work on explanation dialogue \cite{walton2011dialogue}, grounding it in a well-known argumentation framework \cite{prakken2006formal}. In addition, they provide formalisms of \emph{commitment stores} and \emph{understanding stores} for maintaining what each party in the dialogue is committed to, and what they already understand.  This is necessary to prevent circular arguments. They further define how to shift between different dialogues in order to enable nested explanations, in which an explanation produces a new why--question, but also to shift from an explanation to an argumentation dialogue, which supports nested argument due to a challenge from an explainee, as noted by \citeA{antaki1992explaining}. The rules define when this dialectical shift can happen, when it can return to the explanation, and what the transfer of states is between these; that is, how the explanation state is updated after a nested argument dialogue.
 
\subsection{Social Explanation and XAI}

This section presents some ideas on how research from social explanation  affects researchers and practitioners in XAI.

\subsubsection{Conversational Model}

The conversational model of explanation according to \citeA{hilton1990conversational}, and its subsequent extension by \citeA{antaki1992explaining}  to consider argumentation, are appealing and useful models for explanation in AI. In particular, they are appealing because of its generality --- they can be used to explain human or agent actions, emotions, physical events, algorithmic decisions, etc. It abstracts away from the cognitive processes of causal attribution and explanation selection, and therefore does not commit to any particular model of decision making, of how causes are determined, how explanations are selected, or even any particular mode of interaction. 

One may argue that in digital systems, many explanations would be better done in a visual manner, rather than a conversational manner. However, the models of  \citeA{hilton1990conversational}, \citeA{antaki1992explaining}, and \citeA{walton2011dialogue} are all independent of language. They define interactions based on questions and answers, but these need not be verbal. Questions could be asked by interacting with a visual object, and answers could similarly be provided in a visual way.  While Grice's maxim are about conversation, they apply just as well to other modes of interaction. For instance, a good visual explanation would display only quality explanations that are relevant and relate to the question --- these are exactly Grice's maxims.

I argue that, if we are to design and implement agents that can truly explain themselves, in many scenarios, the explanation will have to be interactive and adhere to maxims of communication, irrelevant of the media used. For example, what should an explanatory agent do if the explainee does not accept a selected explanation?

\subsubsection{Dialogue}
\label{sec:social-explanation:XAI:dialogue}

\citeauthor{walton2011dialogue}'s  explanation dialogues \cite{walton2004new,walton2007dialogical,walton2011dialogue}, which  build on  well-accepted models from argumentation, are closer to the notion of computational models than that of \citeA{hilton1990conversational} or  \citeA{antaki1992explaining}. While \citeauthor{walton2011dialogue} also abstracts away from the cognitive processes of causal attribution and explanation selection, his dialogues are more idealised ways of how explanation can occur, and thus make certain assumptions that may be reasonable for a model, but of course, do not account for all possible interactions. However, this is appealing from an explainable AI perspective because it is clear that the interactions between an explanatory agent and an explainee will need to be scoped to be computationally tractable. \citeauthor{walton2011dialogue}'s models provide a nice step towards implementing  \citeauthor{hilton1990conversational}'s conversational model.

 \citeauthor{arioua2015formalizing}'s \emph{formal} model for explanation \cite{arioua2015formalizing} not only brings us one step closer to a computational model, but also nicely brings together the models of \citeA{hilton1990conversational} and \citeA{antaki1992explaining} for allowing arguments over claims in explanations. Such formal models of explanation could work together with concepts such as \emph{conversation policies} \cite{greaves2000conversation} to implement explanations.

\begin{changed}
The idea of interactive dialogue XAI is not new. In particular, a body of work by \citeauthor{cawsey1993planning} \cite{cawsey1991generating,cawsey1992explanation,cawsey1993planning} describes EDGE: a system that generates natural-language dialogues for explaining complex principles. \citeauthor{cawsey1993planning}'s work was novel because it was the first to investigate discourse within an explanation, rather than discourse more generally. Due to the complexity of explanation, \citeauthor{cawsey1993planning} advocates \emph{context-specific},  \emph{incremental} explanation, interleaving planning and execution of an explanation dialogue. EDGE separates content planning (what to say) from dialogue planning (organisation of the interaction). Interruptions attract their own sub-dialog. The flow of the dialogue is context dependent, in which context is given by: (1) the current state of the discourse relative to the goal/sub-goal hierarchy; (2) the current \emph{focus} of the explanation, such as which components of a device are currently under discussion; and (3) assumptions about the user's knowledge. Both the content and dialogue are influenced by the context. The dialogue is planned using a rule-based system that break explanatory goals into sub-goals and utterances. Evaluation of EDGE \cite{cawsey1993planning} is anecdotal, based on a small set of people, and with no formal evaluation or comparison. 

At a similar time, \citeA{moore1993planning} devised a system for explanatory text generation within dialogues that also considers context. They explicitly reject the notion that \emph{schemata} can be used to generate explanations, because they are too rigid and lack the intentional structure to recover from failures or misunderstandings in the dialogue. Like \citeauthor{cawsey1993planning}'s EDGE system, \citeauthor{moore1993planning}  explicitly represent the user's knowledge, and plan dialogues incrementally. The two primary differences from EDGE is that \citeauthor{moore1993planning}'s system explicitly models the effects that utterances can have on the hearer's mental state, providing flexibility that allows recovery from failure and misunderstanding; and that the EDGE system follows an extended explanatory plan, including probing questions, which are deemed less appropriate in \citeauthor{moore1993planning}'s application area of advisory dialogues.
The focus of \citeauthor{cawsey1993planning}'s and \citeauthor{moore1993planning}'s work are in applications such as intelligent tutoring, rather than on AI that explains itself, but many of the lessons and ideas generalise.

EDGE and other related research on interactive explanation considers only verbal dialogue. As noted above, abstract models of dialogue such as those proposed by \citeA{walton2011dialogue} may serve as a good starting point for multi-model interactive explanations.
\end{changed}

\subsubsection{Theory of Mind}
\label{sec:social-explanation:model-of-other}

In Section~\ref{sec:philosophical-foundations:model-of-self}, I argue that an  explanation-friendly \emph{model of self} is required to provide meaningful explanations. However, for social explanation, a \emph{Theory of Mind} is also required. Clearly, as part of a dialog, an explanatory agent should at least keep track of what has already been explained, which is a simple model of other and forms part of the explanatory context. However, if an intelligent agent is operating with a human explainee in a particular environment, it could may have access to more complete models of other, such as the other's capabilities and their current beliefs or knowledge; and even the explainee's model of the explanatory agent itself. If it has such a model, the explanatory agent can exploit this by tailoring the explanation to the human observer. \citeA{halpern2005causes-part-II} already considers a simplified idea of this in their model of explanation, but other work on epistemic reasoning and planning \cite{fagin1995reasoning,muise2015aaai} and planning for interactive dialogue \cite{petrick2016using} can play a part here. These techniques will be made more powerful if they are aligned with user modelling techniques used in HCI \cite{fischer2001user}.

\begin{changed}
While the idea of Theory of Mind in AI is not new; see for example \cite{von2017minds,dignum2014autistic}; it's application to explanation has not been adequately explored. Early work on XAI took the idea of dialogue and user modelling seriously. For example, \citeauthor{cawsey1993planning}'s EDGE system, described in Section~\ref{sec:social-explanation:XAI:dialogue}, contains a specific user model to provide better context for interactive explanations \cite{Cawsey1993UserMI}. \citeauthor{Cawsey1993UserMI} argues that the user model must be integrated closely with explanation model to provide more natural dialogue. The EDGE user model consists of two parts: (1) the \emph{knowledge} that the user has about a phenomenon; and (2) their `level of expertise'; both of which can be updated during the dialogue. EDGE uses dialogues questions to build a user model, either explicitly, using questions such as ``Do you known X?'' or ``What is the value of Y?'', or implicitly, such as when a user asks for clarification.  EDGE tries to guess other indirect knowledge using logical inference from this direct knowledge. This knowledge is then used to tailor explanation to the specific person, which is an example of using epistemic relevance to select explanations.  \citeauthor{cawsey1993planning} was not the first to consider user knowledge; for example, \citeauthor{weiner1980blah}'s BLAH system \cite{weiner1980blah} for incremental explanation also had a simple user model for knowledge that is used to tailor explanation, and \citeauthor{weiner1980blah} refers to Grice's maxim of quality to justify this. 

More recently, \citeA{chakraborti2017explanation} discuss preliminary work in this area for explaining plans. Their problem definition consists of two planning models: the explainer and the explainee; and the task is to align the two models by minimising some criteria; for example, the number of changes. This is an example of using epistemic relevance to tailor an explanation.  \citeauthor{chakraborti2017explanation} class this as contrastive explanation, because the explanation contrasts two models. However, this is not the same use of the term `contrastive' as used in social science literature (see Section~\ref{sec:philosophical-foundations:contrastive-explanation}), in which the contrast is an explicit foil provided by the explainee as part of a question.
\end{changed}

\subsubsection{Implicature}

It is clear that in some settings, implicature can play an important role. Reasoning about implications of what the explainee says could support more succinct explanations, but just as importantly, those designing explanatory agents must also keep in mind what people could infer from the literal explanations --- both correctly and incorrectly.

Further to this, as noted by \citeA{dodd1980leading}, people interpret explanations relative to the intent of the explainer. This is important for explainable AI because one of the main goals of explanation is to establish trust of people, and as such, explainees will be aware of this goal. It is clear that we should quite often assume from the outset that trust levels are low. If explainees are sceptical of the decisions made by a system, it is not difficult to imagine that they will also be sceptical of explanations provided, and could interpret explanations as biased.

\subsubsection{Dilution}

Finally, it is important to focus on dilution. As noted in the introduction of this paper, much of the work in explainable AI is focused on causal attributions. The work outlined in Section~\ref{sec:cognitive-processes} shows that this is only part of the problem. While presenting a casual chain may allow an explainee to fill in the gaps of their own knowledge, there is still a likely risk that the less relevant parts of the chain will dilute those parts that are crucial to the particular question asked by the explainee. Thus, this again emphasises the importance of explanation selection and relevance.

\begin{changed}
\subsubsection{Social and Interactive Explanation}

The recent surge in explainable AI has not (yet) truly adopted the concept socially-interactive explanation, at least, relative to the first wave of explainable AI systems such as that by \citeA{Cawsey1993UserMI} and \citeA{moore1993planning}. I hypothesise that this is largely due to the nature of the task being explained. Most recent research is concerned with explainable machine learning, whereas early work explained symbolic models such as expert systems and logic programs. This influences the research in two ways: (1) recent research focuses on how to abstract and simplify uninterpretable models such as neural nets, whereas symbolic approaches are relatively more interpretable and need less abstraction in general; and (2) an interactive explanation is a goal-based endeavour, which lends itself more naturally to symbolic approaches. Given that early work on XAI was to explain symbolic approaches, the authors of such work would have more intuitively seen the link to interaction. Despite this, others in the AI community have recently re-discovered the importance of social interaction for explanation; for example, \cite{weld2018intelligible,shams2016normative}, and have noted that this is a problem that requires collaboration with HCI researchers.
\end{changed}


\section{Conclusions}
\label{sec:conc}

In this paper, I have argued that explainable AI can benefit from existing models of how people define, generate, select, present, and evaluate explanations. I have reviewed what I believe are some of the most relevant and important findings from social science research on human explanation, and have provide some insight into how this work can be used in explainable AI. 

In particular, we should take the four major findings noted in the introduction into account in our explainable AI models: (1) why--questions are contrastive; (2) explanations are selected (in a biased manner); (3) explanations are social; and (4) probabilities are not as important as causal links. I acknowledge that incorporating these ideas are not feasible for all applications, but in many cases, they have the potential to improve explanatory agents. I hope and expect that readers will also find other useful ideas from this survey.

It is clear that adopting this work into explainable AI is not a straightforward step. From a social science viewpoint, these models will need to be refined and extended to provide good explanatory agents, which requires researchers in explainable AI to work closely with researchers from philosophy, psychology, cognitive science, and human-computer interaction. Already, projects of this type are underway, with impressive results; for example, see \cite{kulesza2011oriented,kulesza2015principles,ribeiro2016should}.

\subsubsection*{Acknowledgements}

The author would like to thank Denis Hilton for his review on an earlier draft of this paper, pointers to several pieces of related work, and for his many insightful discussions on the link between explanation in social sciences and artificial intelligence. The author would also like to thank several others for critical input of an earlier draft: Natasha Goss, Michael Winikoff, Gary Klein, Robert Hoffman, and the anonymous reviewers; and  Darryn Reid for his discussions on the link between self, trust, and explanation.

This work was undertaken while the author was on sabbatical at the Universit\'{e} de Toulouse Capitole, and was partially funded by Australian Research Council DP160104083 \emph{Catering for individuals' emotions in technology development} and and a Sponsored Research Collaboration grant from the Commonwealth of Australia Defence Science and Technology Group and the Defence Science Institute, an initiative of
the State Government of Victoria.


\bibliographystyle{elsarticle-num-names-alpha}
\bibliography{references,other}

\end{document}